\documentclass[default,iicol]{sn-jnl}

\usepackage{graphics,amssymb,amsmath,epsfig,color}
\usepackage{graphicx}

\usepackage{graphicx}%
\usepackage{multirow}%
\usepackage{amsmath,amssymb,amsfonts}%
\usepackage{amsthm}%
\usepackage{mathrsfs}%
\usepackage[title]{appendix}%
\usepackage{xcolor}%
\usepackage{textcomp}%
\usepackage{manyfoot}%
\usepackage{booktabs}%
\usepackage{algorithm}%
\usepackage{algorithmicx}%
\usepackage{algpseudocode}%
\usepackage{listings}%
\usepackage{lettrine}

\usepackage{chngpage}

\usepackage[utf8]{inputenc} 
\usepackage[T1]{fontenc}    
\usepackage{hyperref}       
\usepackage{url}            
\usepackage{booktabs}       
\usepackage{amsfonts}       
\usepackage{nicefrac}       
\usepackage{microtype}      
\usepackage{lipsum}
\usepackage{fancyhdr}       
\usepackage{graphicx}       
\graphicspath{{media/}}     
\usepackage{subcaption}
\usepackage{amsmath}

\usepackage{multibib}
\definecolor{myorange}{RGB}{252,244,228}

\begin{document}
\title[Article Title]{Kernel-Elastic Autoencoder for Molecular Design}


\author{\fnm{Haote} \sur{Li}}\email{haote.li@yale.edu}

\author{\fnm{Yu} \sur{Shee}}\email{yu.shee@yale.edu}

\author{\fnm{Brandon} \sur{Allen}}\email{brandon.allen@yale.edu}

\author{\fnm{Federica} \sur{Maschietto}}\email{federica.maschietto@yale.edu}

\author{\fnm{Victor} 
\sur{Batista}}\email{victor.batista@yale.edu}

\affil{\orgdiv{Department of Chemistry}, \orgname{Yale University}, \city{New Haven}, \state{Connecticut} \postcode{06520}, \country{USA}}



\abstract{We introduce the Kernel-Elastic Autoencoder (KAE), a self-supervised generative model based on the transformer architecture with enhanced performance for molecular design. KAE employs two innovative loss functions: modified maximum mean discrepancy (m-MMD) and weighted reconstruction $(\mathcal{L}_{WCEL})$. The m-MMD loss has significantly improved the generative performance of KAE when compared to using the traditional KL loss of VAE, or standard MMD. Including the weighted reconstruction loss $\mathcal{L}_{WCEL}$, KAE achieves valid generation and accurate reconstruction at the same time, allowing for generative behavior that is intermediate between VAE and AE not available in existing generative approaches. Further advancements in KAE include its integration with conditional generation, setting a new state-of-the-art benchmark in constrained optimizations. Moreover, KAE has demonstrated its capability to generate molecules with favorable binding affinities in docking applications, as evidenced by AutoDock Vina and Glide scores, outperforming all existing candidates from the training dataset. Beyond molecular design, KAE holds promise to solve problems by generation across a broad spectrum of applications.}

\maketitle

\section{Introduction}\label{sec1}

The advent of generative models has precipitated a revolutionary shift in the development of methods for drug discovery, revealing new opportunities to swiftly identify ideal candidates for specific applications~\cite{maziarka2020mol, moret2020generative, skalic2019shape, wang2021multi, bengio2021gflownet, hoogeboom2022equivariant,prykhodko2019novo}. The Variational Autoencoder (VAE) model has emerged amongst these models as an approach with extraordinary capabilities that can be adapted for molecule generation via character, grammar, and graph-based representations~\cite{kingma2013auto, gomez2018automatic, kusner2017grammar, jin2018junction}.

Autoencoders (AEs) encode the input data by compression into a low-dimensional space~\cite{ballard1987modular}. Though providing a high lower bound for accurate reconstruction, such space is not well-structured and in some regions, the decoder does not generate output that resembles the training data, thereby limiting its generative capabilities. Sacrificing reconstruction performance, VAEs mitigate this disadvantage by enforcing encoded latent vectors to known prior distributions. Upon decoding samples from those distributions, VAEs generate outputs mimicking the training data. An outstanding challenge of great interest to drug discovery is to harness the power of VAEs to generate molecular candidates with optimal properties during the screening phase of molecular development, while preserving AE's high reconstruction rate for precise lead candidate optimizations.

Generative models are typically evaluated for molecule generation using novelty (N), uniqueness (U), validity (V), and reconstruction (R) metrics. NUVR metric, which is the product of them, captures the trade-off between these four factors,  the so-called NUV to R tradeoff, as a model with high reconstruction ability usually does not achieve high metrics for novelty, uniqueness, and validity.

Optimizing the design of molecules near a reference molecule requires robust reconstruction, as proximity in latent space should correlate with proximity in the value of the desired property. Accurate reconstruction also allows for interpolation between molecular motifs with intermediate properties between promising lead compounds~\cite{hoffman2022optimizing,van2001property,he2021molecular,chen2021deep}.

\begin{figure*}[!h]
    \centering
    \includegraphics[width=0.6\textwidth]{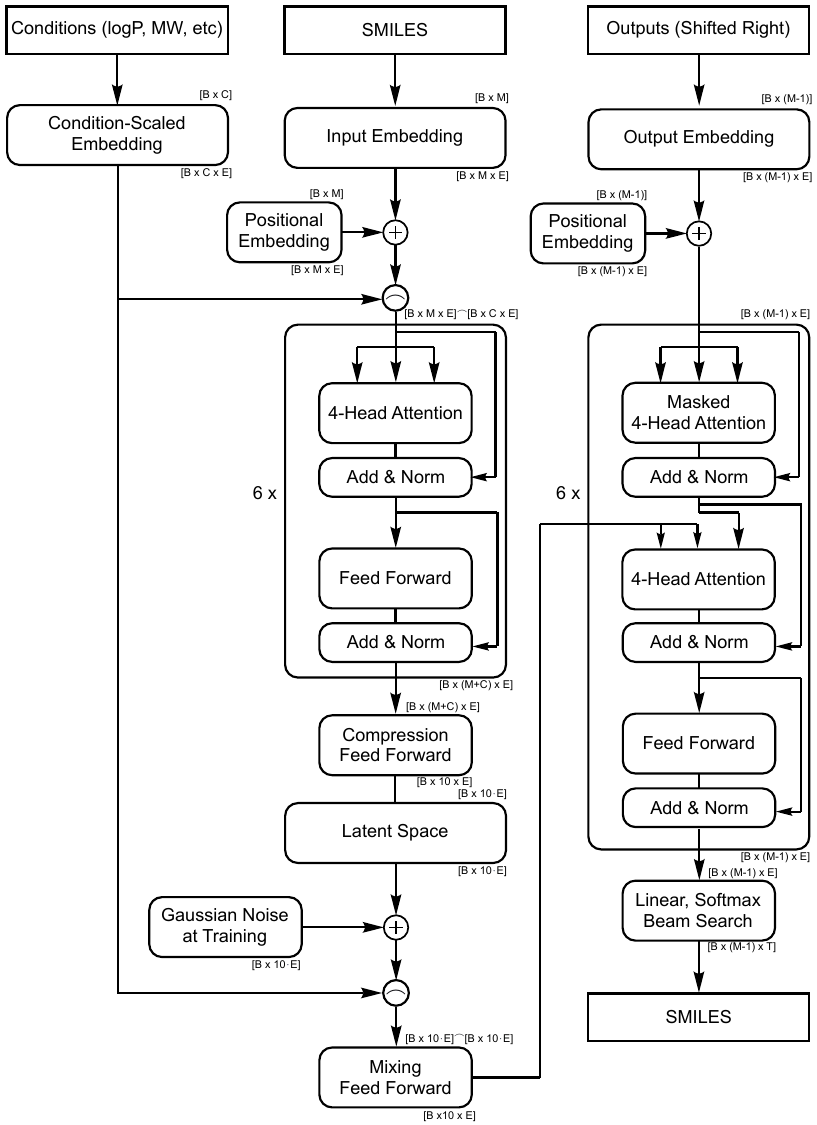}
    \caption{\textbf{Conditional KAE transformer architecture.} KAE consists of 6 encoder layers, 6 decoder layers, and a latent space for conditional generations. During training, the condition is concatenated after positional embedding and provided as input to the 4-head attention encoder. The condition is also concatenated with the latent vector before a mixing layer. During training, Gaussian noise is added to the latent vectors.  The decoder output is then passed through a linear layer and softmax function, producing the probabilities of output tokens for each character in the dictionary of size $T$.}
    \label{fig:architecture}
\end{figure*}

{\em Kernel-Elastic Autoencoder} (KAE) stands out as a new category for a self-supervised generative model based on a modified maximum mean discrepancy and weighted reconstruction loss functions. Leveraged by the Transformer architecture\cite{vaswani2017attention,dollar2021attention, jiang2020transformer, wang2019t}, KAE (Figure ~\ref{fig:architecture}) effectively overcomes the NUV-R tradeoff by combining the merits of both autoencoder (AE) and variational autoencoder (VAE) models. KAE's loss function is a modified version of the Maximum Mean Discrepancy (MMD), inspired by~\cite{zhao2019infovae, ucar2019bridging,louizos2015variational}, that shapes the latent space and enables better performance than using Kullback-Leibler (KL) divergence loss used in VAEs.  When coupled to the weighted cross-entropy loss ($\mathcal{L}_{WCEL}$), KAE, without any checking for molecular grammar or chemical rules, outperforms both string and graphical-based models in generation tasks while exhibiting nearly flawless reconstruction, as demonstrated on the ZINC250k testing sets. The freedom to adjust KAE's behavior through the $\mathcal{L}_{WCEL}$ gives the "Elastic" term in the naming.

When implemented to solve optimization problems, KAE outperforms the state-of-the-art by a substantial 28\%~\cite{richards2022conditional}. Additionally, KAE tackles the problem of molecular docking by finding suitable binding ligands with conditional generation, as demonstrated using the dataset from GFlowNet~\cite{bengio2021gflownet}. Superior candidates from the baseline and the training data are independently verified by both Autodock Vina~\cite{trott2010autodock} and Glide~\cite{halgren2004glide,friesner2006extra}, demonstrating its efficacy and practicality.

\section{Result}\label{sec2}

\subsection{KAE Performance}
The overall performance of the KAE (Figure ~\ref{fig:architecture}) compared to state-of-the-art generative models is shown in Table~\ref{table:NUVR_comparison}. As described in the Methods section, KAE combines a modified-MMD (m-MMD) loss and the Weighted Cross Entropy Loss ($\mathcal{L}_{WCEL}$), with hyperparameters $\lambda$ and $\delta$, and exhibits the generative capabilities of VAEs as well as the exact reconstruction objectives of AEs. 
\begin{table*}
  \centering
  \caption{\textbf{Comparison of performance of molecular generative models trained with the ZINC250K dataset.} Assessment of the capabilities of the models to generate novel (N), unique (U), valid (V), and properly reconstructed (R) molecules. Validity (V w/o) indicates that the generated strings have not been post-processed using chemical knowledge to enforce corrections. NUV results were obtained from averaging over 5 iterations of sampling 10,000 random vectors from latent space while the reconstruction rate was calculated using all molecules from the testing dataset. The two KAE models in the table were trained using loss functions with $\lambda$ = 1 and 3.5 and $\delta$ = -1 and 1. The choice of $\delta$ = -1 is a special case of $\mathcal{L}_{WCEL}$ and is equivalent to not using any AE objectives. The validity check selects alternative candidates from beam search.}

\colorbox{myorange}{
\centering 
  \begin{tabular}{lccccccc}
    \toprule
    \cmidrule(r){1-2}
    Method  & N  & U & V w/o& V & NUV & R & NUVR\\
    \midrule
    CVAE~\cite{gomez2018automatic}$^a$ & 0.980  & 0.021 & 0.007 & N/A  &0.0001 & 0.446 & 5e-6\\
    GVAE~\cite{kusner2017grammar}$^a$ & 1.000  & 1.000 & 0.072 & N/A  &0.072 & 0.537 & 0.039\\
    JTVAE~\cite{jin2018junction}$^a$ & 1.000  & 1.000 & 0.935 & 1.000  &0.935 & 0.767 & 0.717\\
    MoFlow~\cite{zang2020moflow} & 1.000  & 0.999 & 0.818 & 1.000  &0.817 & 1.000$^b$ & 0.817\\
    Rebalanced~\cite{yan2022molecule} & 1.000  & 1.000 & 0.907 & 0.938  &0.907 & 0.927 & 0.841\\
    GraphDF~\cite{luo2021graphdf} & 1.000  & 0.992 & 0.890 & 1.000  &0.883 & 1.000$^b$ & 0.883\\
    ALL SMILES~\cite{alperstein2019all}$^a$ & 1.000  & 1.000 & N/A & 0.985  &N/A & 0.874 & N/A\\
    $\beta$-VAE~\cite{richards2022conditional} & 0.998  & 0.983 & 0.983 & 0.988  &0.964 & N/A & 
    N/A\\
    
    \hline
    \textbf{KAE ($\lambda = 1, \delta = -1$)}    & 0.998 & 0.994 & 0.863 & N/A & 0.856   & 0.992 & 0.849   \\
    
    \textbf{KAE ($\lambda = 3.5, \delta = 1$)} & \textbf{0.996}  & \textbf{0.973} & \textbf{0.997} & \textbf{1.000}  & \textbf{0.966} & \textbf{0.997} & \textbf{0.963}\\   
    
    \bottomrule
  \end{tabular}
}
\begin{tablenotes}
\centering

\item[a] $^a$Results obtained from sampling 1,000 vectors from latent space.
\item[b] $^b$Reconstruction rates were obtained on training datasets.
\end{tablenotes}

\label{table:NUVR_comparison}
\end{table*}
KAE was evaluated according to the fraction of generated molecules that are novel (N), unique (U), and valid (V). A molecule is considered novel if it is not included in the training dataset. Uniqueness is defined as the absence of duplicates in the set of generated molecules. A molecule is counted as valid if its SMILES representation is syntactically correct and passes the RDKit chemical semantics checks~\cite{landrum2016rdkit}. Additionally, reconstruction (R) is successful if and only if the decoder regenerates the input SMILES sequence matching every single character.

Maximum validity and reconstruction was achieved by using the  Weighted Cross-Entropy Loss $\mathcal{L}_{WCEL}(\lambda, \delta)$ defined by Eq.~(\ref{eq:WCEL}) where the hyperparameter $\delta$ controls the AE-like objective (see S.I. for a discussion of the effect of changing $\lambda$ and $\delta$). The best results for the NUVR metric were obtained by using a combination of $\lambda = 3.5$ and $\delta = 1$. 

With the same Transformer architecture, KAE is compared to approaches using different loss functions (SI, Figure ~\ref{fig:KL_loss_comparison}) by assessing the validity and reconstruction. KAE trained with Gaussian noise added to latent space vectors exhibited the highest percentage of valid SMILES strings, while models trained with the KL divergence exhibited much lower validity and significantly slower improvements for validity during training. The analysis of novelty and uniqueness showed that models with noise (i.e., with Gaussian noise added to latent space vectors) performed much better than the corresponding models without noise when trained with the standard MMD (s-MMD) or modified MMD (m-MMD, see method section). Additionally, the NUV metric showed that models trained with m-MMD outperformed models trained with s-MMD.

\subsection{Learning Behavior}

\begin{figure*}[!h]
    \centering
    \captionsetup[subfigure]{position=top, labelfont=bf, textfont=normalfont, singlelinecheck=off, justification=raggedright}
    
    \begin{subfigure}{.475\textwidth}
      \caption{}
      \centering
      \includegraphics[width=\textwidth]{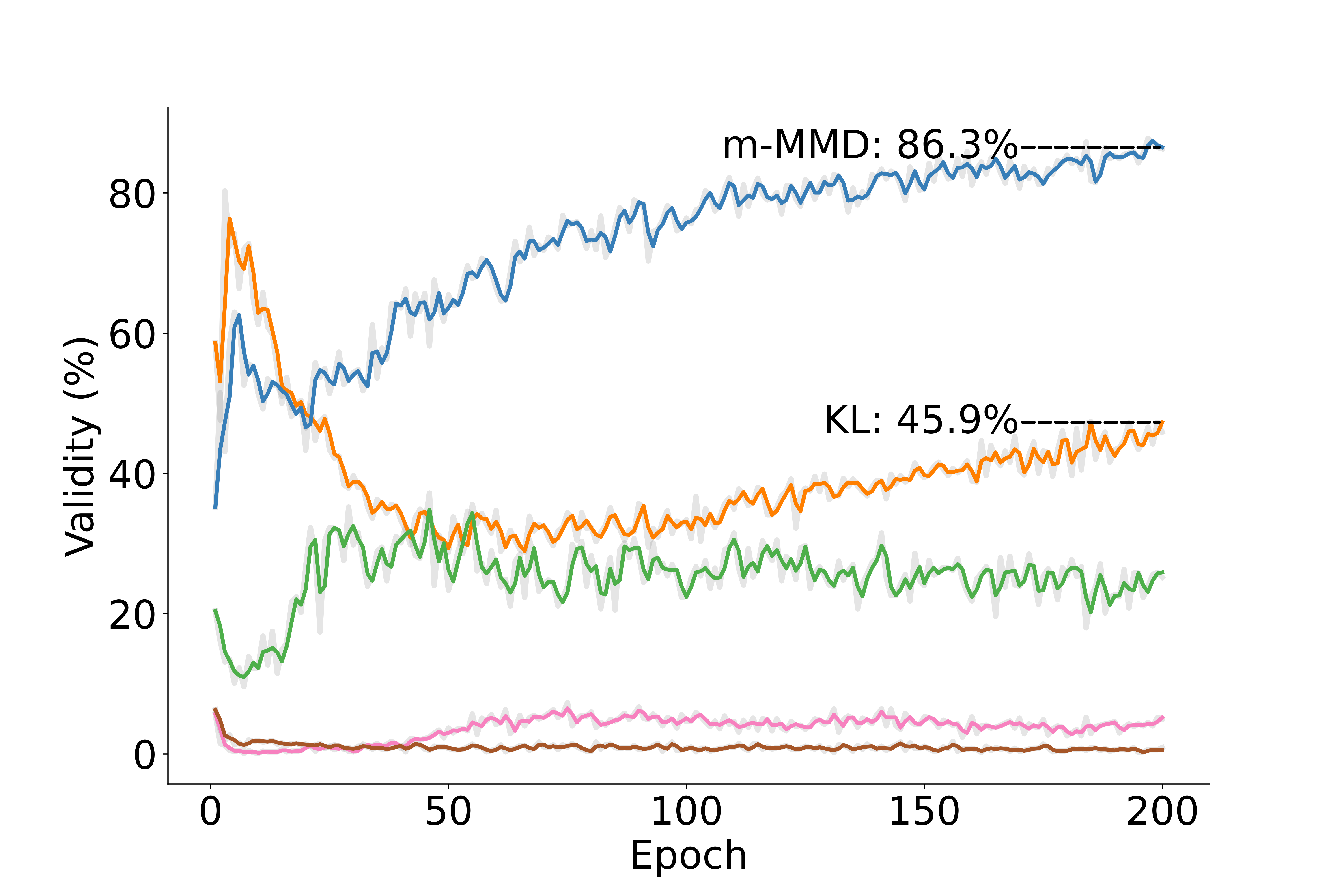} 
      \label{fig:validity_comparison}
    \end{subfigure}
    \begin{subfigure}{.475\textwidth}
      \caption{}
      \centering
      \includegraphics[width=\textwidth]{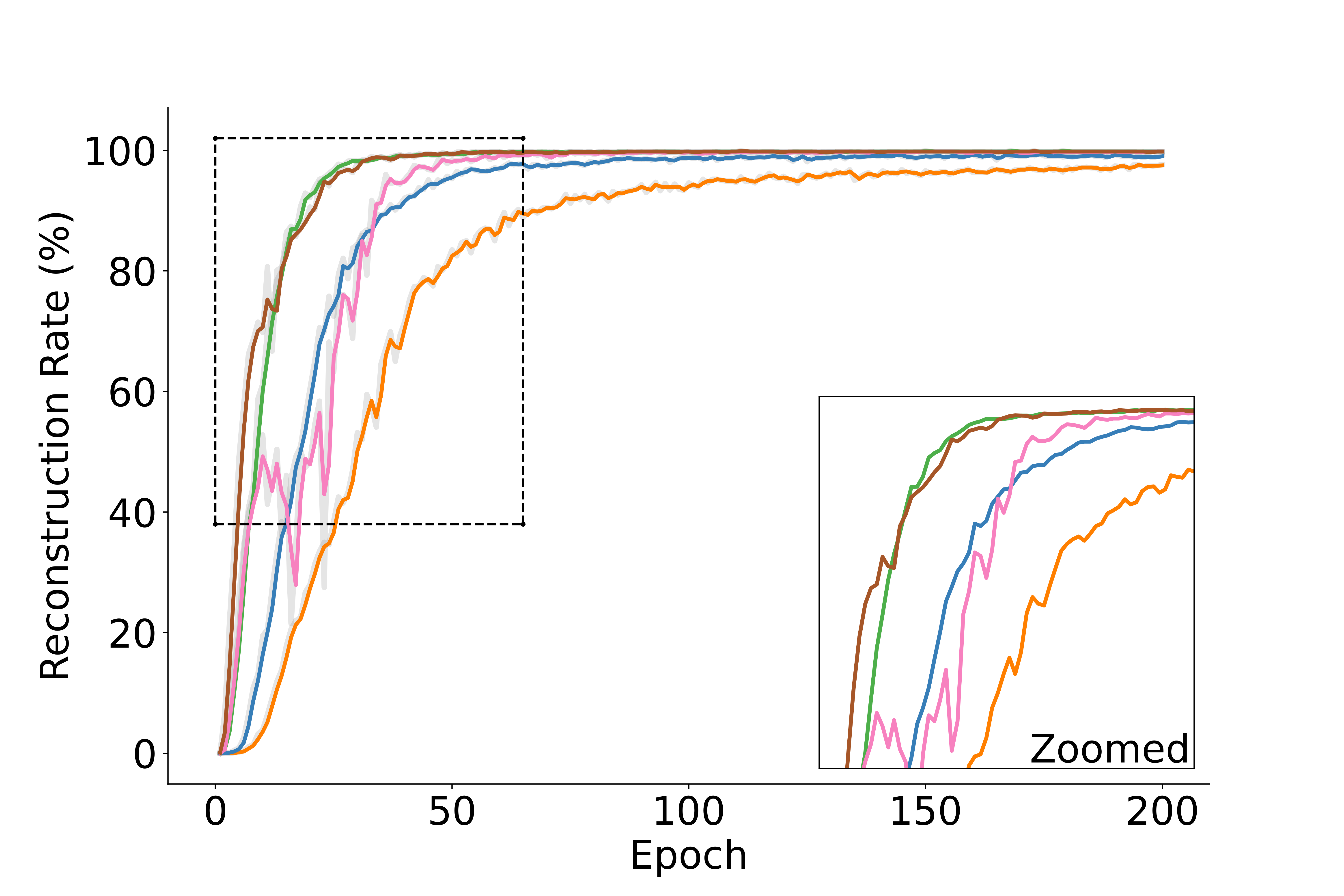} 
      \label{fig:reconstruction_comparison}

    \end{subfigure}
    \begin{subfigure}{.475\textwidth}
      \caption{}
      \centering
      \includegraphics[width=\textwidth]{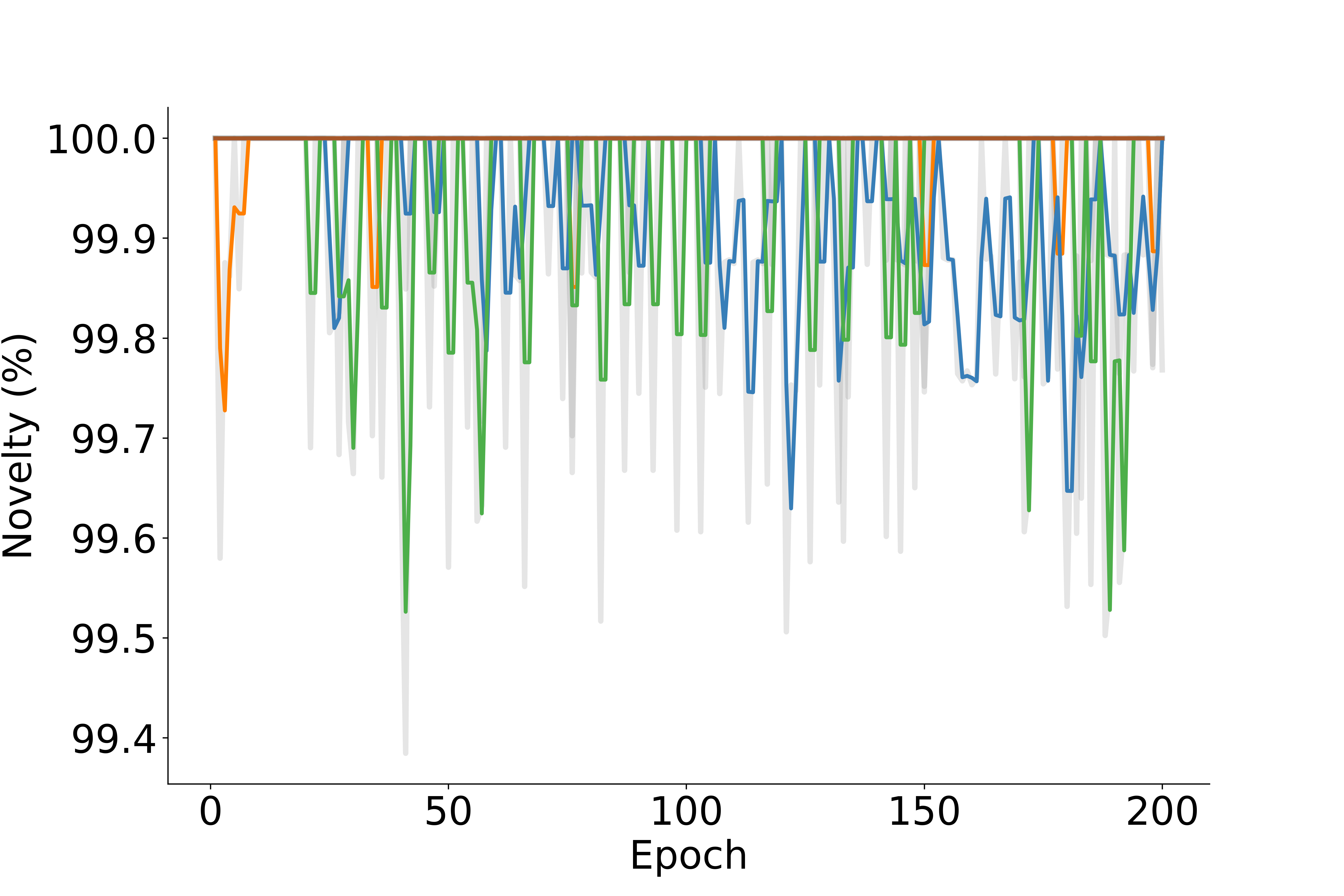} 
      \label{fig:novelty_comparison}
    \end{subfigure}
    \begin{subfigure}{.475\textwidth}
      \caption{}
      \centering
      \includegraphics[width=\textwidth]{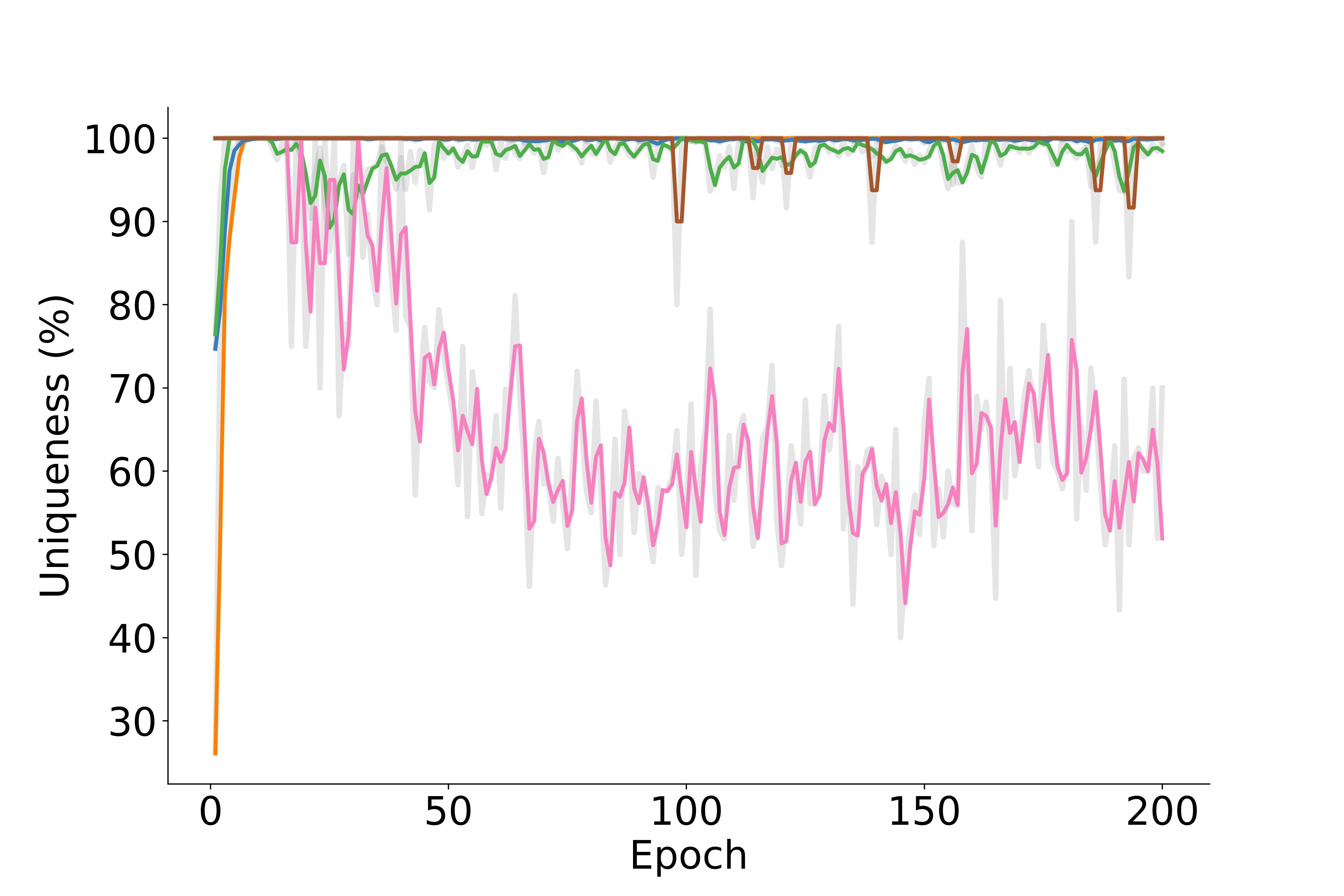} 
      \label{fig:uniqueness_comparison}
    \end{subfigure}
    
    \hfill
    \begin{subfigure}{\textwidth}
      \centering
      \includegraphics[width=0.9\textwidth]{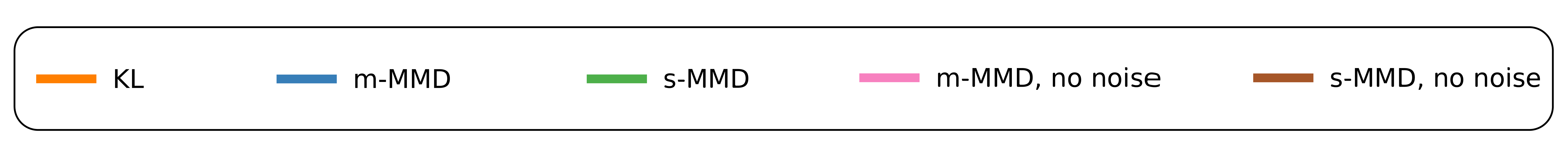} 
    \end{subfigure}
    \hfill

\caption{\textbf{Comparison of learning rates for models trained with m-MMD loss, s-MMD loss, and KL divergence loss.} (a) Validity evaluated at each epoch. (b) Fraction of molecules properly reconstructed as a function of epochs. (c) Novelty evaluated at each epoch. (d) The uniqueness at each epoch. The model labeled as KL includes an extra layer that estimates the standard deviation of each latent vector. The models labeled with m-MMD are trained with the loss $\mathcal{L}_{CEL} + \textit{m-MMD}(\lambda=1)$, s-MMD with $\mathcal{L}_{CEL} + \textit{s-MMD}(\lambda=1)$, and KL with $\mathcal{L}_{VAE}=\mathcal{L}_{CEL} + \textit{KL}(\lambda=1)$. "No noise" means no noise is added to the latent vectors during training. }

\label{fig:KL_loss_comparison}
\end{figure*}

We have analyzed the KAE behavior by comparing under the same architecture but with various loss functions (Figure ~\ref{fig:KL_loss_comparison}). 
The reconstruction was evaluated from 1000 molecules from the validation set at every epoch. Figure~\ref{fig:KL_loss_comparison} shows the improvement in validity, uniqueness, novelty, and reconstruction along the training process for models based on a loss that combines the weighted cross-entropy $\mathcal{L}_{WCEL}$($\lambda$,$\delta$), defined by Eq.~(\ref{eq:WCEL}), with m-MMD (m-MMD($\lambda$), Eq.~\ref{eq:m-mmd}), s-MMD, Eq.~\ref{eq:s-mmd}), or KL divergence. All models were trained with the ZINC250K dataset for 200 epochs, with $\lambda=1$ and $\delta=-1$. When $\lambda = -\delta$, the weighted cross-entropy loss ($\mathcal{L}_{WCEL}$, Eq.~\ref{eq:WCEL}) reduces to the standard cross-entropy loss ($\mathcal{L}_{CEL}$). Additionally, we examine the effect of noise while training with the m-MMD loss. The results (Figure ~\ref{fig:KL_loss_comparison}) indicate that the KAE model using m-MMD loss with Gaussian noise added to the latent space exhibits the best performance. The models exhibit significant differences in their ability to generate valid SMILES strings and reconstruct input molecules. The m-MMD model trained with noise in latent space generated the highest percentage of valid SMILES strings, making it preferable to other models. For example, the model trained with KL divergence exhibited much lower validity and a significantly slower learning rate. The assessment of novelty and uniqueness also shows that s-MMD and m-MMD models trained with Gaussian noise added in latent space (noisy models) performed much better than the corresponding models without noise. The reason for adding noise is to prevent the model from remembering exactly the locations of the latent vectors and the decoder has to see the multitude of possible outcomes related to the region of the decoding latent vector. Further, the decision to add a Gaussian noise on top of confining the latent vector to the same Gaussian through m-MMD is to maximize the overlap of the distribution of all latent vectors with respect to the distribution of any individual latent vector. This approach is different from VAE as the VAE has the option to output small variances for some latent vectors which could reduce the probability of sampling corresponding instances from its prior distribution.

\subsection{Conditional KAE}
In this section, the performance of the Conditional-KAE (CKAE) (Figure ~\ref{fig:architecture}) on the constraint optimization task is investigated.
\begin{table*}
 \caption{\textbf{Comparison of performance of various conditional generative models.} The table presents the average PLogP improvements computed for the set of 800 lowest ranking molecules from the ZINC250K dataset, as well as the mean Tanimoto similarities of the best candidate molecules compared to their respective starting molecules (standard deviations reported after $\pm$). The success rate indicates the percentage of molecules for which the algorithm successfully achieved modifications resulting in higher PLogP values within the specified similarity constraint. The ZINC250K result corresponds to the highest PLogP improvement obtained by searching within the ZINC250K dataset itself. Our approach outperforms the search against the training data and demonstrates the highest performance when combining our model with the SES method.}
  \centering
\colorbox{myorange}{
  \begin{tabular}{lcccc}
    \toprule
    \cmidrule(r){1-2}
    Method$^a$     & PLogP-Improvement     & Tanimoto Similarity & Success Rate \\
    \midrule
    
    JT-VAE~\cite{jin2018junction} & 0.84 $\pm$ 1.45  & 0.51 $\pm$ 0.1 & 83.6\%  \\
    MHG-VAE~\cite{kajino2019molecular} & 1.00 $\pm$ 1.87  & 0.52 $\pm$ 0.11 & 43.5\%    \\
    GCPN~\cite{you2018graph}& 2.49 $\pm$ 1.30  & 0.47 $\pm$ 0.08 & 100\%    \\
   
    Mol-CycleGAN~\cite{maziarka2020mol} & 2.89 $\pm$ 2.08  & 0.52 $\pm$ 0.10 & 58.75\%    \\
    MolDQboot~\cite{zhou2019optimization} & 3.37 $\pm$ 1.62  & N/A & 100\%    \\
    
    \textbf{ZINC250K (This work)} & 4.64 $\pm$ 2.33  & 0.48 $\pm$ 0.16 & 97.88\%     \\
    MoFlow~\cite{zang2020moflow} & 4.71 $\pm$ 4.55  & \textbf{0.61 $\pm$ 0.18} & 85.75\%       \\
    \textbf{Random Sample (This work)} & 4.78 $\pm$ 2.08  & 0.43 $\pm$ 0.03 & 81.75\%     \\

    MNCE-RL~\cite{xu2020reinforced} & 5.29 $\pm$ 1.58  & 0.45 $\pm$ 0.05 & 100\%     \\
    $\beta$-VAE~\cite{richards2022conditional}& 5.67 $\pm$ 2.05  & 0.42 $\pm$ 0.05 & 98.25\%      \\
    \hline
    \textbf{CKAE (This work)} & \textbf{7.67 $\pm$ 1.61}  & 0.42 $\pm$ 0.02 & \textbf{100\%}     \\
    \bottomrule
  \end{tabular}
  }
\label{table:Constrained_Optimization}
\begin{tablenotes}
\centering
\item[a]{$^a$Tanimoto similarity constraint = 0.4}
\end{tablenotes}
\end{table*}
CKAE generates new candidates conditioned on properties such as PLogP or docking scores. Here, we demonstrate the capabilities of CKAE as applied to the PLogP values defined, as follows:~\cite{gomez2018automatic, jin2018junction}:
\begin{equation}
    PLogP(m) = LogP(m) - SA(m) - \text{{ring}}(m),
    \label{eq:plogp}
\end{equation}
where $LogP$ is the octanol-water partition coefficient of molecule $m$ calculated using Crippen's approach from the atom contributions~\cite{wildman1999prediction}. SA is the synthetic accessibility score,~\cite{ertl2009estimation} while $\text{{ring}}(m)$ corresponds to the number of rings with more than six members in the molecule.

\begin{figure}[!h]
    \centering
    \includegraphics[width=0.45\textwidth]{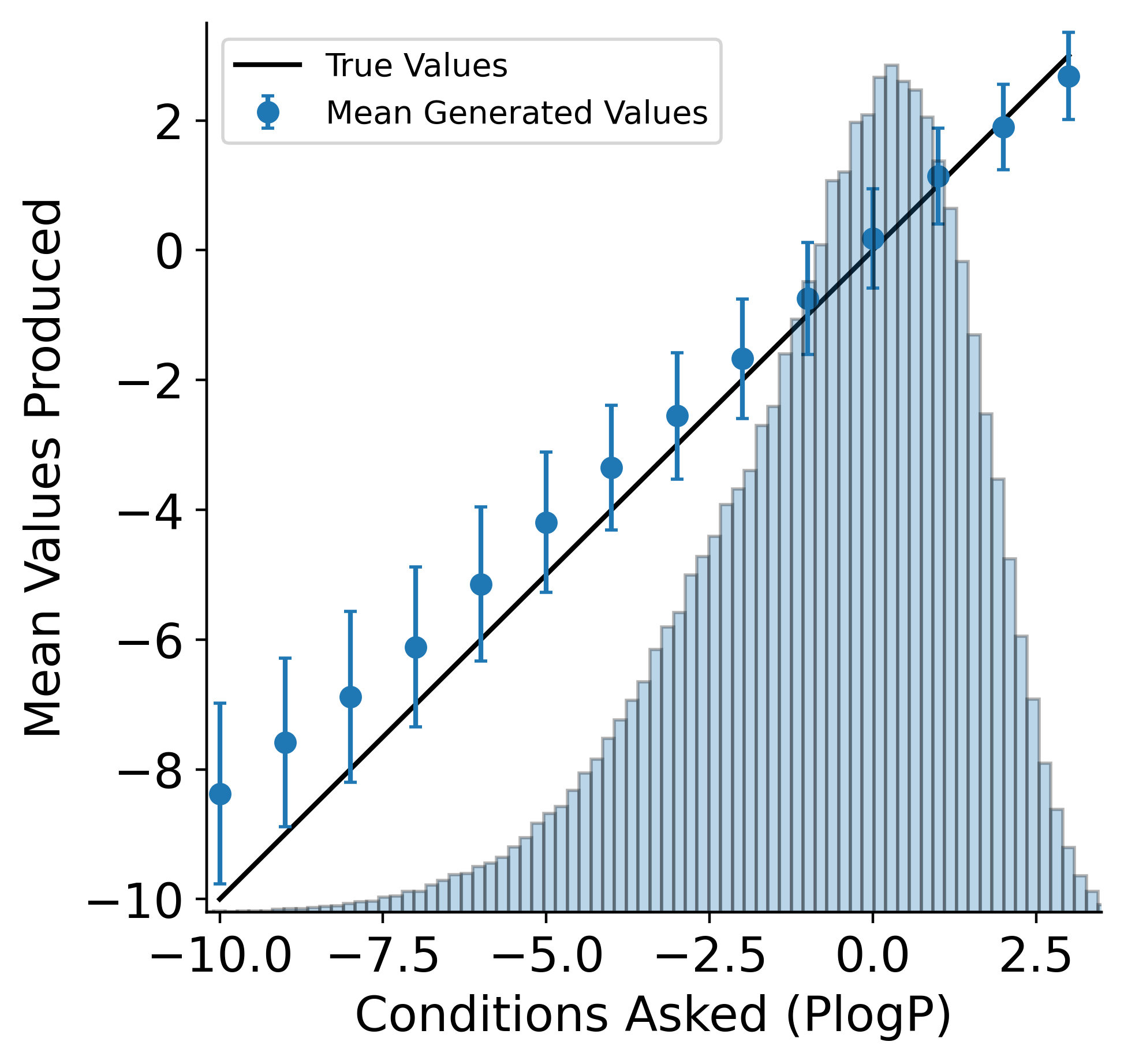}
    \caption{\textbf{CKAE correlation performance.} The blue dots represent the mean PLogP values of 1,000 molecules generated by CKAE, as a function of the condition PLogP value. The error bars on each dot indicates the associated standard deviation as estimations of errors. The black line shows the ground truth values strongly correlated with the mean PLogP values. The histogram shows the underlying distribution of the training dataset over the entire range of PLogP values.}
    \label{fig:conditions_corr}
\end{figure}

To demonstrate that CKAE generates molecules that are strongly correlated to the conditioned value, we analyzed the correlation between the properties of CKAE-generated molecules and the specified input condition. Figure~\ref{fig:conditions_corr} shows the mean PLogP value obtained from 1,000 CKAE-generated molecules, strongly correlated to the PLogP value used as a condition (correlation coefficient 0.9997). The distribution of PLogP values of the training set, rendered as a histogram in Figure~\ref{fig:conditions_corr}, shows the range of PLogP values used for CKAE training. We have also trained a separate model using the dataset from Lim et al \cite{lim2018molecular} who developed a Conditional-VAE (CVAE) with Recurrent Neural Network (RNN) architectures to sample molecules given five distinct pharmaceutically relevant properties. The comparison between CKAE and CVAE from Lim et al further shows that CKAE generates candidates correlating to the asked conditions and outperforms the given baseline by a wide margin (~Table \ref{table:Lim_Compare}).

\begin{table*}[h] 

  \caption{\textbf{Performance of CKAE compared to CVAE, as applied to conditional molecular generation} We impose strict counting criteria for the CKAE statistics so that only valid, novel, and unique molecules are considered attempts. Therefore, the number of valid molecules will be equal to the number of attempts. To further compare to the CVAE method by Lim et al where there is more than one valid molecule per attempt, we have applied beam search with a beam size of 10 (labeled CKAE w. Beam). When beam search is used, the number of valid molecules reports the number of valid, novel, and unique candidates derived from all attempts. The success rate is defined as 100 times the rate of finding a candidate within a 10\% error range of each property per attempt. The result shows CKAE, without using a method to derive more candidates per output SMILES strings is better than CVAE with RNN architectures. Further, if beam search is applied, CKAE significantly outperforms the given baseline.}
  \centering
  
  \colorbox{myorange}{
  \begin{tabular}{ccccc}
    \toprule
    
    Method& Target     & Attempts   & Number of Valid Molecules & Success Rate (\%)\\
    \midrule
    
     CVAE Lim et al& Aspirin & 28,840 & 32,567 & 0.34  \\
     CVAE Lim et al& Tamiflu & 15,960 & 34,696  & 0.62 \\
     
       \hline
       
     \textbf{CKAE} & Aspirin & \textbf{4743}& 4743& \textbf{2.11}\\  
     \textbf{CKAE} & Tamiflu & \textbf{3715}& 3715& \textbf{2.63}\\

     \textbf{CKAE w. Beam}& Aspirin & \textbf{671}& 4221& \textbf{14.90}\\  
     \textbf{CKAE w. Beam}& Tamiflu & \textbf{436}& 3927& \textbf{22.94}\\  
     
    \bottomrule
  \end{tabular}
  }
  
  \label{table:Lim_Compare}
 
\end{table*}

Instead of using regressors to navigate in the latent space\cite{jin2018junction,richards2022conditional,ma2021gf,yan2022molecule}, a procedure called Similarity Exhaustion Search (SES) was developed for constraint optimizations. SES aims to find molecules that are both similar to the target molecule and have higher desired properties (e.g., PLogP) by using the same or slightly perturbed latent vector representations with gradually increasing conditions. Formally, $f(z,c) \approx f(z + \Delta_{z}, c + \Delta_{c})$ for small values of $\Delta_{z}$ and $\Delta_{c}$ where $f(z,c)$ is the decoding output function of latent vector $z$ subject to the condition $c$ (e.g., $\text{PLogP}=c$). When the generative model has high enough NUVR values, it is able to pinpoint the exact latent vector location and perform an exhaustive search for all possible $\Delta z$. Therefore, SES combines beam search with iterative sampling under various conditions to identify chemically similar molecules that closely resemble the target compound in the latent space. The details of SES can be found in Section~\ref{sec:SES}. 

Table~\ref{table:Constrained_Optimization} shows 1. the results of optimizing the 800 lowest PLogP-valued molecules from the ZINC250K dataset to generate similar molecules (Tanimoto similarity < 0.4) with larger PLogP values.~\cite{zhou2019optimization}; 2. The mean difference in PLogP values; 3. The Tanimoto similarity between the best candidate molecules and their starting molecules for each method. The success rate measures the percentage of molecules that achieved modifications with higher PLogP values within their similarity constraints.

Additionally, CKAE performance was assessed as compared to direct search from the ZINC250K training set. For each of the 800 molecules, its similarity value with respect to all other 250K entries was calculated, and the compound with the highest PLogP value that remained within the 0.4 Tanimoto similarity constraint was identified. This particular outcome is labeled ``ZINC250K'' in Table~\ref{table:Constrained_Optimization}.

We further compared CKAE to direct search using randomly sampled latent vectors with different conditions (PLogP values from -10 to 10 scanned with a step size of 0.1). At each step, instead of using encoder-provided latent vectors. 800 vectors were randomly sampled from the latent space and decoded using beam search with a beam size of 15. The outcomes of this search are marked as ``Random Search'' in Table~\ref{table:Constrained_Optimization}.

\subsection{CKAE for Ligand Docking}

\subsubsection{Comparison to GFlowNet}
Table~\ref{table:Diverse_Search} shows the performance of the CKAE model as applied to the generation of small molecule inhibitors that bind to the active site of the enzyme soluble epoxide hydrolase (sEH), as compared to results obtained with GFlowNet for the same active site~\cite{bengio2021flow,bengio2021gflownet}.
\begin{table*}[h] 
  \caption{\textbf{Performance of the CKAE model on molecular docking as compared to GFlowNet.} Top 10, 100, and 1000 rewards are the averages of the docking scores of molecules generated at the corresponding thresholds. The Top-1000 similarity is the mean of all pair-wise similarities. Lower similarity between generated molecules indicates greater diversity, which is desirable. For docking, the higher rewards are better.}
  \centering
  \colorbox{myorange}{
  \begin{tabular}{ccccc}
    \toprule
    \cmidrule(r){1-2}
    Method     & Top 10 reward   & Top 100 reward & Top 1000 reward &Top-1000 similarity\\
    \midrule
    
     GFlowNet & 8.36  & 8.21 & 7.98 & 0.44 \\
     Training Data & 9.62   & 8.78 & 7.86  & 0.58 \\
       \hline
     \textbf{CKAE (This work)} & \textbf{11.15}   & \textbf{10.46} & \textbf{9.63}  & 0.63 \\  
    \bottomrule
  \end{tabular}
  }
  
  \label{table:Diverse_Search}
 
\end{table*}

CKAE was trained using the same dataset of 300,000 molecules used for training GFlowNet~\cite{Bengio2022}, each with a binding energy calculated using AutoDock~\cite{trott2010autodock} (see Section~\ref{subsec:AutodockMethods}). Binding energies were converted to a reward metric, using a custom scaling function.  
Results in Table~\ref{table:Diverse_Search} correspond to the mean reward for the top 10, 100 and 1000 best scoring molecules from a pool of $10^{6}$ NUV molecules generated by the CKAE model. Rewards were computed from the Autodock Vina binding scores. Average Tanimoto similarities were computed using a Morgan Fingerprint with a radius of 2. 

Table~\ref{table:Diverse_Search} shows that CKAE achieves similar performance to GFlowNet in molecular docking, and generates molecules with higher rewards at the top 10, 100, and 1000 thresholds, without significantly sacrificing the similarity score. In fact, CKAE was able to generate molecules scoring as high as 11.45, which exceeds the maximum reward of 10.72 in the training database. This demonstrates the capabilities of CKAE for generative extrapolation, which allows for applications to generative dataset augmentation including molecules with scoring values beyond the range of the original dataset.


\subsubsection{Glide Analysis}

\begin{figure*}[!ht]
    \centering
    \captionsetup[subfigure]{singlelinecheck=off, justification=raggedright, labelfont=bf, position=top}
    \begin{subfigure}{.95\textwidth}
      \caption{}
      \centering
      \includegraphics[width=\textwidth]{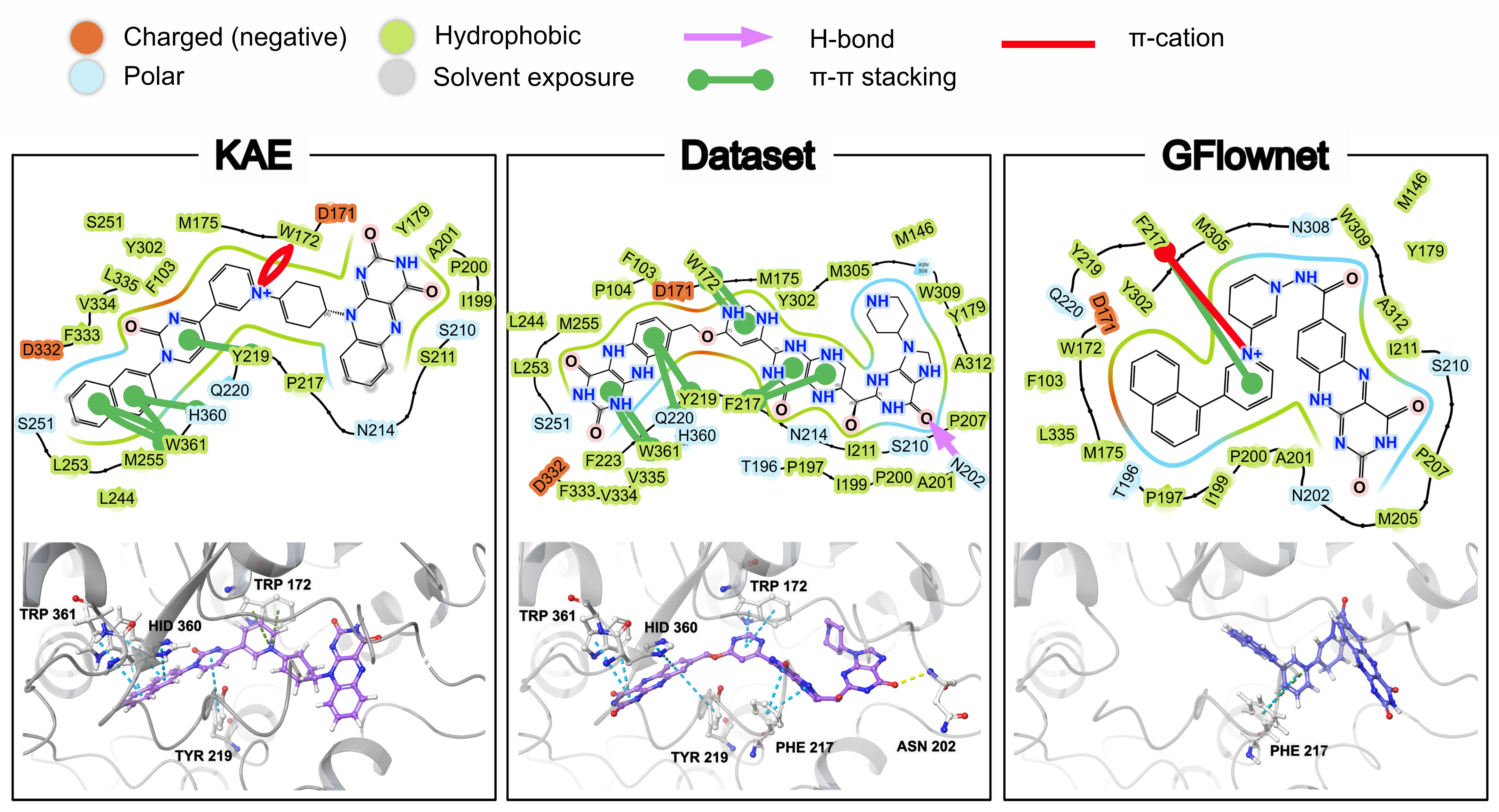} 
      \label{fig:KAE-Ligand-Interaction-Figure-New}
    \end{subfigure}
    \hfill
    \begin{subfigure}{0.95\textwidth}
      \centering
      \includegraphics[width=\textwidth]{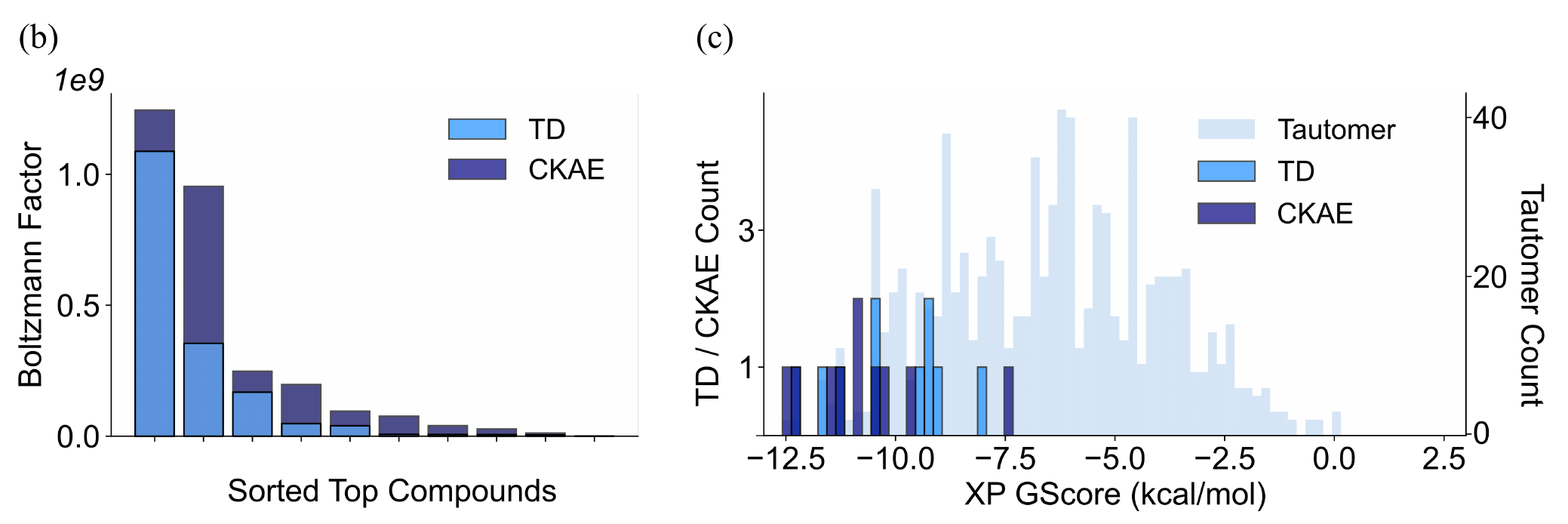} 
    \end{subfigure}

    \begin{subfigure}{0pt} 
        \phantomcaption
        \label{fig:top_bzm}
    \end{subfigure}
    \begin{subfigure}{0pt} 
        \phantomcaption
        \label{fig:tauto}
    \end{subfigure}
    \hfill
    \caption{\textbf{Glide analysis of molecular inhibitors docked at the active site of sEH.} \textbf{(a)} Binding interactions of top scoring molecules generated by CKAE (left), searched from the training dataset (middle), and generated by GFlowNet (right). \textbf{(b)} Extra Precision (XP) Glide score Boltzmann factors for the top ten candidates obtained from the CKAE and training dataset (TD) show that the top ranking CKAE-generated outperform the top molecules from the TD ensemble. \textbf{(c)} Histogram of Glide XP docking scores, showing that top scoring inhibitors generated by CKAE or TD outperform 869 tautomers generated from the top ten candidates of the two datasets.}
\end{figure*}

A comparison of the ligand-receptor interactions established by the top-scoring CKAE, training dataset (TD) and GFlowNet candidates, respectively is shown in Figure~\ref{fig:KAE-Ligand-Interaction-Figure-New}.  KAE’s top candidate exhibits superior docking performance compared to top-scoring candidates in both the training dataset and GFlowNet. In terms of fitting within the pocket, the top CKAE candidate occupies a substantially larger volume within the receptor binding region when compared to the other two. The improved fit is also evidenced by the broader array of stabilizing interactions. These interactions include a series of $\pi$-$\pi$ stacking and $\pi$-cation interactions. In addition to occupying the pocket entirely, the CKAE-generated molecules are devoid of unfavorable clashes, further underscoring the effectiveness of the model in generating effective candidates in the context of molecular docking.

Figure~\ref{fig:top_bzm} shows the analysis of the best scoring molecules generated by CKAE and direct search from the training dataset (TD), as assessed by the Glide molecular docking program that is an integral part of the Schrödinger Suite of software~\cite{halgren2004glide, friesner2006extra}. Figure~\ref{fig:top_bzm} thus provides an independent assessment of the quality of the best-scoring CKAE-generated molecules, showing that CKAE-generated molecules outperform the TD counterparts in terms of ranking as determined by the nature of the interactions established at the binding site.

The docking procedure employed an identical receptor grid size as used for Autodock Vina~\cite{trott2010autodock} calculations, and the candidates sourced from both the training dataset and CKAE, were docked onto the same receptor structure, using the highest scoring pose derived from Autodock Vina~\cite{trott2010autodock} calculations, as described in Section~\ref{subsec:GlideMethods}.


A dataset comprised of 869 tautomers was curated with high structural similarity, including the top ten CKAE-derived molecules and the top ten TD molecules, as well as a set of tautomers of the same molecules generated by changing protonation and enantiomeric states
to analyze the quality of the top-performing hits relative molecular tautomers (molecules with different arrangements of atoms and bond). The results shown in Figure ~\ref{fig:tauto} revealed that the top-ranking candidates from both CKAE and TD outperformed other contenders (tautomers) when compared against the dataset of tautomers. These results confirmed that the highest-scoring molecular structures obtained from CKAE and TD remained superior, even when compared to a large number of structurally similar alternatives, confirming the reliability and quality of molecules generated by CKAE.

As examined by both Autodock Vina~\cite{trott2010autodock} and Glide~\cite{halgren2004glide,friesner2006extra}, it is clear that CKAE generated molecules that bind better to the active site of sEH than those of the training dataset. The generated higher-scoring molecules can then be used for dataset augmentation, for retraining purposes, allowing the model to generate even higher-scoring molecules.

\section{Discussion}

KAE allows the integration of the strengths of both VAE and AE frameworks in applications to molecular design. The KAE loss, with hyperparameters $\lambda$ and $\delta$, controls varying degrees of VAE and AE features as needed for the specific applications. 

In the context of molecule generation, KAE outperformed VAE approaches in terms of generation validity without the need for additional chemical knowledge-based checks, while achieving reconstruction performance near 100\% accuracy akin to the AE.  With beam search decoding ~\cite{graves2012sequence,bengio2013advances,guo2023beam} multiple candidates per latent vector can be derived. This enriched KAE's generation diversity and validity. In the context of conditional generation, CKAE generates molecules that exhibit excellent correlation with the input condition, including molecules with a desired property (e.g., specific value of PLogP, or reward value upon docking to a specific binding site of a biological target). 

In the constrained optimization task, the CKAE model exhibits significant improvements, with an average increase of 7.67 $\pm$ 1.61 PLogP units. CKAE achieves a 100\% success rate, indicating that modifications leading to higher PLogP values were successfully achieved for all molecules within the defined similarity constraints.  This improvement surpasses directly searching from the training dataset by over 65\%. The comparison to ``Random Search''  shows the strength of KAE's accurate reconstruction which makes searching around the molecules much more efficient. 
 
Using Glide~\cite{halgren2004glide,friesner2006extra}, the validation for CKAE's generated high binding affinity candidates reveals that they consistently outperform those from the training dataset as well as all structurally similar tautomers, demonstrating CKAE's ability of extrapolation and the quality of the generated molecules.

As demonstrated in this article, KAE can be used to tackle docking problems with binding affinity and constrained optimization with PlogP. Similar but unlimited to the context of molecule designs, KAE can be effectively employed to address a wide spectrum of problems, especially for those that are labeled as sample-property pairs.

\section{Methods}\label{sec:methods}

\subsection{Model Architecture}
KAE treats molecule generation as a natural language processing task. Phrases in the ``source language'' (i.e., SMILES strings) are encoded and compressed into latent vectors and then decoded into the target output with corresponding labels. Major components for KAE include the encoder, compression layer, mixing layer, and decoder.

Source and target masks are created with specified padding tokens to ensure that the encoder and decoder do not attend to padding tokens during training. The SMILES tokens are separately passed through embedding layers of the encoder and decoder to become vectors of size 128. They are then added to the corresponding position embeddings of the same dimensions. Different from the original Transformer implementation that uses fixed sinusoidal functions in the representation, in this work, each positional token's embedding is learned and updated during training.

The input is encoded by the Transformer encoder and compressed into latent space. The compression layer is a single linear layer that applies to the sequence length dimensions. This layer takes in a dimension $M$, the maximum sequence length in the relevant dataset and outputs a dimension of 10. In the case of ZINC250k without using conditions, $M$ is 113 dimensional. The resulting latent tensor therefore has dimensions of 10$\times E$ where $E$ is the embedding size of 128. The latent vectors are then added with noise from a standard Gaussian distribution.
In the CKAE variant, the conditions (i.e., molecule properties) are attached with additional embeddings. Condition-multiplied embeddings are concatenated with the input of the encoder and the latent representation along the sequence length dimension. This allows the model to generate molecules by either interpolating or extrapolating with a desired condition value. The mixing layer is a linear layer that takes in the compressed tensors with the size (10 + number of conditions)$\times E$ and maps them back to 10$\times E$ dimensions. These tensors are treated as the new encoder output which the decoder attends to without encoder masks.
Each decoder layer attends to the encoder outputs through encoder-decoder multi-head attention operations. The decoder outputs are contracted by a linear layer along the embedding dimension, producing a $T$-dimensional vector per token, where $T$ is the dictionary size. This $T$-dimensional vector is then softmaxed, resulting in a probability distribution ($P$) for each possible character ($c$).

\subsection{KAE Loss}\label{sec:KAE_Loss}

The KAE loss function is defined, as follows:
\begin{equation} \label{eq:final}
\mathcal{L}\left(\lambda, \delta \right) = \mathcal{L}_{WCEL}(\lambda, \delta) + \textit{m-MMD}(\lambda),
\end{equation}
where $\textit{m-MMD}(\lambda)$ is a modified version of the regularizing MMD loss, discussed in Sec.~\ref{sec:m-mmd}, and $\mathcal{L}_{WCEL}$ is a weighted cross-entropy loss (\textit{$\mathcal{L}_{WCEL}$}) obtained from outputs generated by decoding the latent vector with and without Gaussian noise added to the latent vector. Based on the original definition of the cross-entropy loss (\textit{CEL}):
\begin{equation}\label{eq:CEL}
\mathcal{L}_{CEL} = -\sum_{s} \sum_{c} Y_{s,c} \log\left( P_{s,c} \right),
\end{equation}
where $P_{s,c}$ is the predicted softmax probability of token $c$ at sequence position $s$ and
$Y_{s,c}$ is the ground truth label equal to one if the token belongs to class $c$ at position $s$, or zero otherwise. Accordingly, we define $\mathcal{L}_{WCEL}$, as follows:
\begin{align}\label{eq:WCEL}
\begin{split}
\mathcal{L}_{WCEL}(\lambda, \delta) &= \frac{-1}{\lambda + \delta + 1}\left[\sum_{s} \sum_{c} Y_{s,c}  \log\left( P_{s,c} \right)\right. \\
&\left.+ (\lambda + \delta)\sum_{s} \sum_{c} Y_{s,c}  \log\left( P_{s,c}^{*} \right)\right].
\end{split}
\end{align}
where $P_{s,c}$ and $P_{s,c}^{*}$ are the predicted softmax values obtained upon decoding the latent vector with and without added Gaussian noise, respectively.

The hyperparameters $\lambda$ and $\delta$ control the significance of the second term in the r.h.s. of Eq.~(\ref{eq:WCEL}) (AE behavior) as well as the relative weight between the m-MMD term and the weighted cross-entropy loss, according to Eq.~(\ref{eq:final}). The function of $\lambda$ is analogous to the $\beta$ parameter in $\beta$-VAE \cite{higgins2017beta}. By adjusting $\lambda$ and $\delta$, the learning objective can be positioned between the VAE and AE objectives. At the extremes, the objective becomes VAE-like (or AE-like) upon weighting more the term with (or without) noise. For example, when $\lambda=1$ and $\delta = -1$, $\mathcal{L}$ is like the VAE loss except that we use m-MMD instead of the KL-divergence.  For AE-like behavior, we choose $\lambda = 0$ and $\delta = 1$. 

The inclusion of $\lambda$ in the second term of Eq.~(\ref{eq:WCEL}) allows larger $\lambda$ values to restrict the latent vectors closer together, penalized by the m-MMD loss. This effect increases the probability of sampling valid latent vectors but reduces distinctions between vectors. Further details on the effect of $\lambda$  in Section~\ref{section:lambda}. The normalization factor of $\frac{1}{\lambda + \delta + 1}$ is derived on the basis to make a linear interpolation between the $\mathcal{L}_{CEL}$ with and without noise.

During training, both the latent vector and the decoder outputs with and without noise are necessary for the calculation of the KAE loss. The latent vectors are penalized based on their differences from 1000 randomly sampled Gaussian vectors ($\vec{G_i}$) using kernel-based metrics\cite{zhao2019infovae}. During training, a noise vector $\epsilon \in \mathbb{R}^D$, with $D$ the dimension of the latent space, is added to the latent vector before passing it to the decoder. The noise vector is generated from a Gaussian normal distribution $\mathcal{N}(\mu,\sigma^2)$ with zero mean $\mu=0$ and unit variance $\sigma=1$.

The center of the Figure~\ref{fig:architecture} captures the training procedure, where information from the latent space is passed to the decoder twice.

One pass resembles an AE-like behavior without noise, while the other pass resembles a VAE-like behavior with added noise to the latent vector before decoding. The reconstructions are both penalized by $\mathcal{L}_{WCEL}$. The parameter $\lambda$ controls the shape of the latent vector distribution and the relative weights between the MMD term and the cross-entropy loss. The parameter $\delta$ controls the relative weights between the AE and VAE objectives.

\subsection{KAE m-MMD Loss}
\label{sec:m-mmd}
The MMD loss 
\cite{gretton2012kernel}, between two distributions having $N_{x}$ and $N_{y}$ samples, is defined as their squared distance calculated in a space $\mathcal{F}$ through the transformer $\phi$:
\begin{equation}
\label{eq:mmd}
\begin{split}
{\text{MMD}}(\vec{x},\vec{y}) &= ||\vec{\mu_{x}}-\vec{\mu_{y}}||^{2}_{\mathcal{F}},\\
&= \vec{\mu_{x}}^{T} \cdot \vec{\mu_{x}} + \vec{\mu_{y}}^{T} \cdot \vec{\mu_{y}} \\&- \vec{\mu_{x}}^{T} \cdot \vec{\mu_{y}} - \vec{\mu_{y}}^{T} \cdot \vec{\mu_{x}},
\end{split}
\end{equation}
where $\vec{\mu_{x}} = \frac{1}{N_{x}}\sum_{i}^{N_{x}} \phi(\vec{x_{i}})$. The space $\mathcal{F}$ is defined by its dot product which can be calculated using a kernel function $\mathcal{K}$.  Introducing the kernel 
\begin{equation}
\mathcal{K}(\vec{x}_i,\vec{y}_j) = \vec{\phi(x_i)}^{T} \cdot \vec{\phi(y_j)},    
\end{equation}
we can write the inner products, as follows:
\begin{equation} \label{eq:inner_product}
    \vec{\mu_{x}}^{T} \cdot \vec{\mu_{y}} = \frac{1}{N_{x}N_{y}}\sum_{i}^{N_{x}}\sum_{j}^{N_{y}} \mathcal{K}(\vec{x}_i,\vec{y}_j),
\end{equation}
so
\begin{align} \label{eq:s-mmdv}
\begin{split}
\text{MMD}(\vec{x},\vec{y}) & = \frac{1}{N_{y}^{2}}\sum_{i}^{N_{y}}\sum_{j}^{N_{y}}\mathcal{K}(\vec{y_{i}},\vec{y_{j}})\\
& + \frac{1}{N_{x}^{2}}\sum_{i}^{N_{x}}\sum_{j}^{N_{x}}\mathcal{K}(\vec{x_{i}},\vec{x_{j}})\\
& - \frac{2}{N_{x}N_{y}}\sum_{i’}^{N_{x}}\sum_{j’}^{N_{y}}\mathcal{K}(\vec{x_{i’}},\vec{y_{j’}}),
\end{split}
\end{align}
where all $\vec{y}$ are sampled from the target Gaussian distribution, and the kernel is defined as follows:
\begin{equation} \label{eq:Kernel}
\mathcal{K}(\vec{\alpha}, \vec{\beta}) = exp(\frac{-\frac{1}{D} \sum_{d=0}^{D} (\alpha_{d} - \beta_{d})^{2}}{2\sigma^2}),
\end{equation}
where $D=10~\times~E$ is the size of the latent dimension and $\sigma=\sqrt{0.32}$ has been empirically chosen (see comparison in \ref{sec:sigma_comparison}).

The first term in the r.h.s. of Eq.~(\ref{eq:s-mmdv}) corresponds to $\vec{\mu_{y}}^{T} \cdot \vec{\mu_{y}}$. It is typically dropped in the loss evaluations since this term does not contribute to the gradients of the loss with respect to the weights during backpropagation. So, the standard-MMD (s-MMD) loss is defined, as follows:
\begin{align} \label{eq:s-mmd}
\begin{split}
\textit{s-MMD}(\lambda) & =\lambda\left[\frac{1}{N_{x}^{2}}\sum_{i}^{N_{x}}\sum_{j}^{N_{x}}\mathcal{K}(\vec{x_{i}},\vec{x_{j}}) \right.\\
&\left. - \frac{2}{N_{x}N_{y}}\sum_{i’}^{N_{x}}\sum_{j’}^{N_{y}}\mathcal{K}(\vec{x_{i’}},\vec{y_{j’}})\right].
\end{split}
\end{align}

For a zero-minimum inner product, the minimum of the first term is achieved at $\vec{\mu_{x}}$ equal zero. So, minimizing the first term promotes all $\phi(\vec{x_{i}})$ to spread out in the space $\mathcal{F}$ while the second term brings $\phi(\vec{x})$ to be similar to the distribution of $\phi(\vec{y})$. 

Based on the s-MMD loss, introduced by Eq.~(\ref{eq:s-mmd}), we define the m-MMD loss, as follows:
\begin{align}
\begin{split}
\textit{m-MMD}(\lambda)=\lambda\left[1 - \frac{1}{N_{x}N_{y}}\sum_{i}^{N_{x}}\sum_{j}^{N_{y}}\mathcal{K}(\vec{x_{i}},\vec{y_{j}})\right] \\
\:
\label{eq:m-mmd}
\end{split}
\end{align}

The constant 1 is added to make m-MMD range from 0 to 1 before the $\lambda$ scaling. A comparative analysis of the effect of using m-MMD versus s-MMD is provided in Supplementary Information (Section~\ref{sec:Latent_Space_And_Performance}).

\subsection{Decoding Methods} \label{sec:decoding_method}
KAE's generation process involves sampling a vector, $\vec{v} \in \mathcal{R}^{10\times E}$ from a $D$-dimensional Gaussian distribution and decoding it. For conditional generation (CKAE), the sampled vector is concatenated with a condition $C$, following its multiplication by its corresponding embedding vector. The resulting vector is subsequently mingled by a fully connected layer, yielding $\vec{L}$ again in $\mathcal{R}^{10\times E}$. The decoder then translates the SMILES string sequence, character by character, with decoder-encoder attention applied to $\vec{L}$.

During decoding, the token ``<SOS>'' is initially supplied. The decoder subsequently generates a probability distribution across $T$ possible tokens for each input. One of the approaches is to continue the predictions using the token possessing the maximum probability, incorporating the token into the next-round input sequence and reiterating the procedure to obtain the next most probable token. This process is repeated until the end-of-sequence token (?) is produced or the sequence length limit is achieved. Besides retaining only the token of highest probability, KAE employed beam search, guided by the hyper-parameter beam size, to derive a broader array of interpretations of the same vector, $\vec{L}$. With a beam size, $B$, where $B \leq T$, a maximum of $B$ outputs are produced from a single decoding procedure.

The beam search algorithm logs the probability of each step for each of the $B$ sequences. For the first step, the top $B$ most probable tokens are selected. In subsequent steps, the model decodes from $B$ input sequences concurrently. Given that each of the $B$ sequences has $T$ potential outcomes for the succeeding token, the total number of potential next-step sequences equates to $B\times T$. These sequences are then ranked according to the sum of their probabilities for all $S$ characters.

In a beam search, the probability of a sequence of tokens indexed from $s, s-1, s-2...$ to $0$ can be represented, as follows: 
\begin{align}
\begin{split}
    &P(s, s-1, s-2,...,0) \\ &= P(s|s-1, s-2,...,0) \times P(s-1, s-2,...,0) \\
    & \:
\end{split}
\end{align}
This can be interpreted as the product of individual probabilities, 
\begin{align} \label{eq:product}
\begin{split}
    &P(s,s-1, s-2,...,0) = P(s|s-1, s-2,...,0) \\ & \times P(s-1|s-2,s-3,...,0)
    \times ... \: P(0)
\end{split}
\end{align}
However, calculations of long sequences based on this equation yield impractically small numbers as every term is smaller than one. Therefore, we sum the log probabilities instead.

For the $B\times T$ sequences with equal sequence length $S$, the probability of the $i^{\text{th}}$ sequence at each position $s$ is denoted as $P_{i,s}$. Excluding the probabilities of padding tokens, the sum of log probabilities, $P_{i}$ for the $i^{\text{th}}$ sequence is computed as:

\begin{equation}
P_{i} = \frac{1}{\sqrt{N_{i}}}\sum_{s\neq pad}^{S} Log(P_{i,s})
\end{equation}

Here, $N_{i}$ represents the quantity of non-padding tokens in sequence $i$.

To foster diversity in decoding, sequence lengths are factored into the computation of $P_{i}$. The $\frac{1}{\sqrt{N_{i}}}$ term counteracts the preference for shorter sequences over longer ones, as longer sequences typically yield smaller sums of log probabilities.

The top $B$ most probable tokens are selected and serve as the inputs for the subsequent iteration, which continues until the maximum sequence length $M$ is attained or all top $B$ candidates have produced the end-of-sequence token, signaling the cessation of decoding.

\subsection{Docking Methods}\label{subsec:AutodockMethods}

The generated molecular structures were evaluated using Autodock Vina~\cite{trott2010autodock}, following a procedure that ensures meaningful comparisons to other molecular generation models, such as GFlowNet~\cite{bengio2021gflownet}. All results were independently tested by using Glide docking from Schrodinger Inc.~\cite{halgren2004glide,friesner2006extra} to ensure the results are robust across different docking software packages.

Autodock Vina is known for its efficiency and speed, making it suitable for high-throughput screening. It employs an empirical scoring function for accurate prediction of binding affinities. On the other hand, Glide utilizes a force field-based scoring function that is widely recognized for its accuracy. In particular, Glide excels at predicting binding poses with high precision and has undergone extensive validation. Its efficacy in handling large and flexible ligands has established it as the gold standard in the field. To ensure meaningful comparisons to GFlowNet~\cite{bengio2021gflownet}, we followed the same procedure implemented for Autodock Vina calculations. Specifically, 20 conformers were used per ligand, exhaustiveness was set to 32, and a maximum of 10 binding modes were generated.

\subsubsection{Glide calculations}\label{subsec:GlideMethods}
The model protein receptor (PDB ID: 4jnc) was prepared by using the adept Protein Preparation Wizard tool in Maestro~\cite{maestro2023}. The protonation states were defined at a neutral pH = 7.0. The protein was subsequently refined via energy minimization using the OPLS4 force field~\cite{lu2021opls}. 

The 3D grid representation of the receptor binding site was prepared by using the Maestro Grid Generation tool, ensuring that the grid size and positioning was perfectly aligned with those used in GFlowNet calculations. All model structures for docking were preared using the LigPrep tool of Maestro. Utilizing the Pre-Dock tool in Maestro, the docked molecules were prepared and assigned charges and protonation states via the OPLS4 force field~\cite{lu2021opls}. The XP (extra precision)~\cite{friesner2006extra,halgren2004glide} flexible docking protocol was then implemented, employing a range of settings designed to optimize the docking accuracy and precision. These included a selection of all predefined functional groups for biased torsional sampling, the addition of Epik state penalties~\cite{greenwood2010epik} to the docking scores~\cite{maestro2023}, and the enhanced planarity setting for conjugated pi groups. 

In the initial step of the docking procedure, 10,000 poses were filtered through the Glide screens, and the top 1,000 poses were selected for energy minimization. The expanded sampling option was utilized to maximize pose flexibility during the search. Ultimately, a single pose was retained for each ligand. The final stage involved refining the best docking poses. Two consecutive refinement steps were performed, each consisting of a post-docking energy minimization on the selected pose, eliminating the need for additional sampling. As a result, highly optimized and reliable docking poses were obtained and compared against those obtained with Autodock Vina~\cite{trott2010autodock} calculations.

\pagebreak
\section*{Acknowledgments}
VSB acknowledges support from the NSF CCI grant (Award No. 2124511) and computational resources from the National Energy Research Scientific Computing Center (NERSC), a U.S. Department of Energy Office of Science User Facility located at Lawrence Berkeley National Laboratory, operated under Contract No. DE-AC02-05CH11231 using NERSC award BES-ERCAP0024372.

\section*{Data Availability}
Pretrained KAE can be accessed through API calls at 
\url{https://demo.ischemist.com/login}. The key to viewing KAE example results is \url{publicdemo}. To use the API, please contact victor.batista@yale.edu.
 
\bibliographystyle{unsrt}  
\bibliography{main}  

\newpage
\onecolumn
\begin{center}
\textbf{\large Supplemental Materials: Kernel-Elastic Autoencoder for Molecular Design}
\end{center}
\setcounter{equation}{0}
\setcounter{figure}{0}
\setcounter{table}{0}
\setcounter{page}{1}
\setcounter{section}{0}

\renewcommand{\theequation}{S\arabic{equation}}
\renewcommand{\thesection}{S\arabic{section}}
\renewcommand{\thefigure}{S\arabic{figure}}
\renewcommand{\thetable}{S\arabic{table}}
\renewcommand{\bibnumfmt}[1]{[S#1]}
\renewcommand{\citenumfont}[1]{S#1}


\section{Training Datasets} \label{training}
The KAE model has been trained on 90\% (225,000) of the entries of the ZINC250K dataset. Within the other split, 1,000 molecules were used for validation and 24,000 were used for testing. In CKAE, the training data included the molecular properties from the ZINC250K library. For the dataset with 300,000 docking candidates from GFlowNet, all entries were used for training.

\section{Data Preparation}
We used the ZINC250K dataset. During dataset preparation, all SMILES strings were first canonicalized and added to the start of sequence tokens ``$\text{!}$" and the end of sequence tokens ``$\text{?}$". The canonicalization gives a unique and unambiguous representation of the molecule. There were 41 unique characters from the database. They were extracted and put into a character-to-token dictionary that allows conversions from characters to tokens. Paddings were added at the end as the 42$^{nd}$ token, making the dictionary size $T$. A token-to-character dictionary was created at the same time for the interpretation of the model output in tokens.
With the character-to-token dictionary, all SMILES representations were converted to the corresponding tokens. Since we use the Transformer architecture, model inputs were made into the same shape for batch training by adding paddings to all sequences. After padding, all sequences have the same length. 
The numerical values of the penalized octanol-water partition coefficient (PLogP) were concatenated to the end of the corresponding tokenized molecules. This adds one extra dimension in the sequence length. The maximum sequence length for each molecule in the dataset is denoted as $M$.
The tokenized dataset is then partitioned into 256-size batches.



\section{Comparison with Scaled KL Divergence}
\begin{figure*}[!h]
    \centering
    \captionsetup[subfigure]{position=top, labelfont=bf, textfont=normalfont, singlelinecheck=off, justification=raggedright}
    
    \begin{subfigure}{.475\textwidth}
      \caption{}
      \centering
      \includegraphics[width=\textwidth]{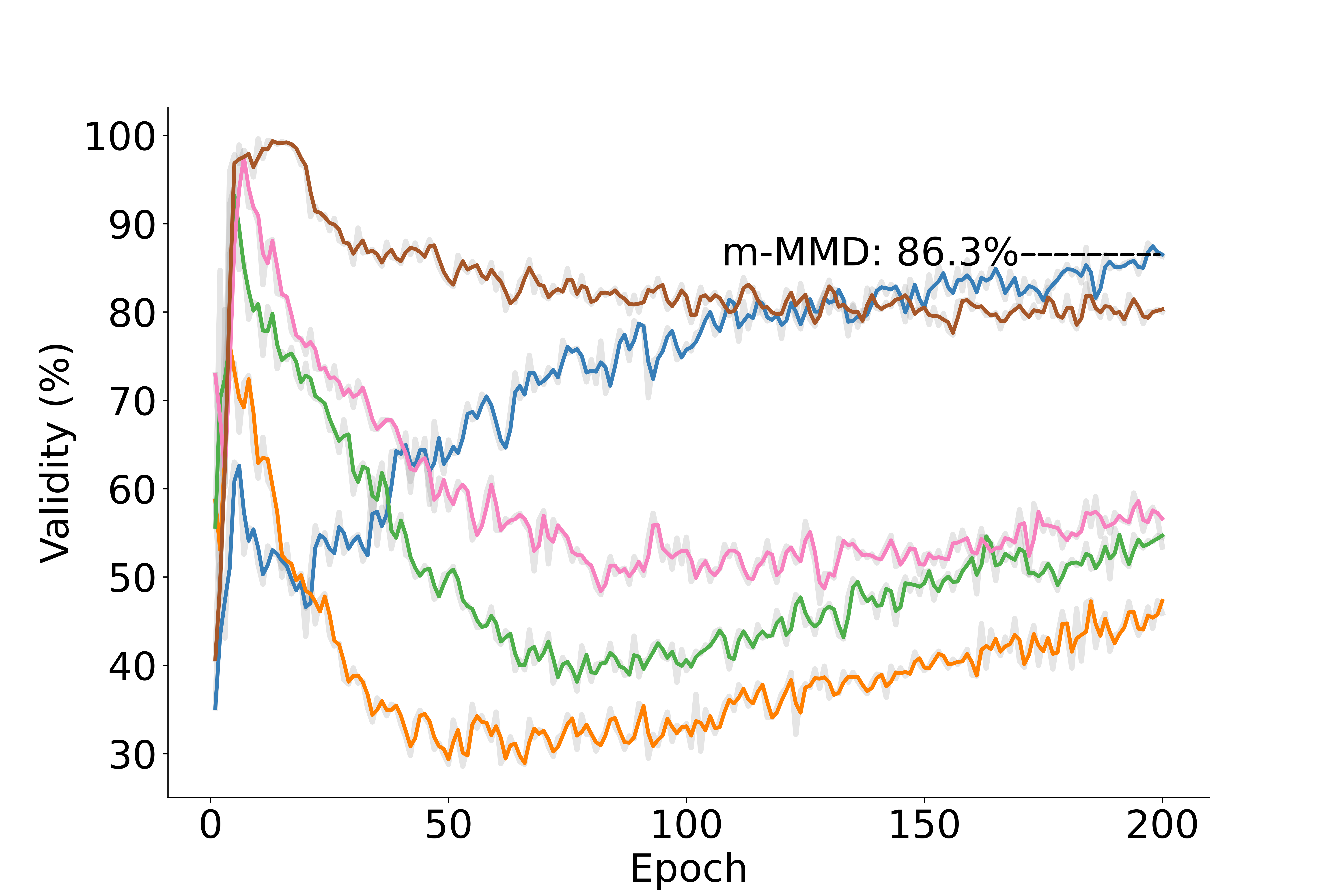} 
      \label{fig:KL_L_validity_comparison}
    \end{subfigure}
    \begin{subfigure}{.475\textwidth}
      \caption{}
      \centering
      \includegraphics[width=\textwidth]{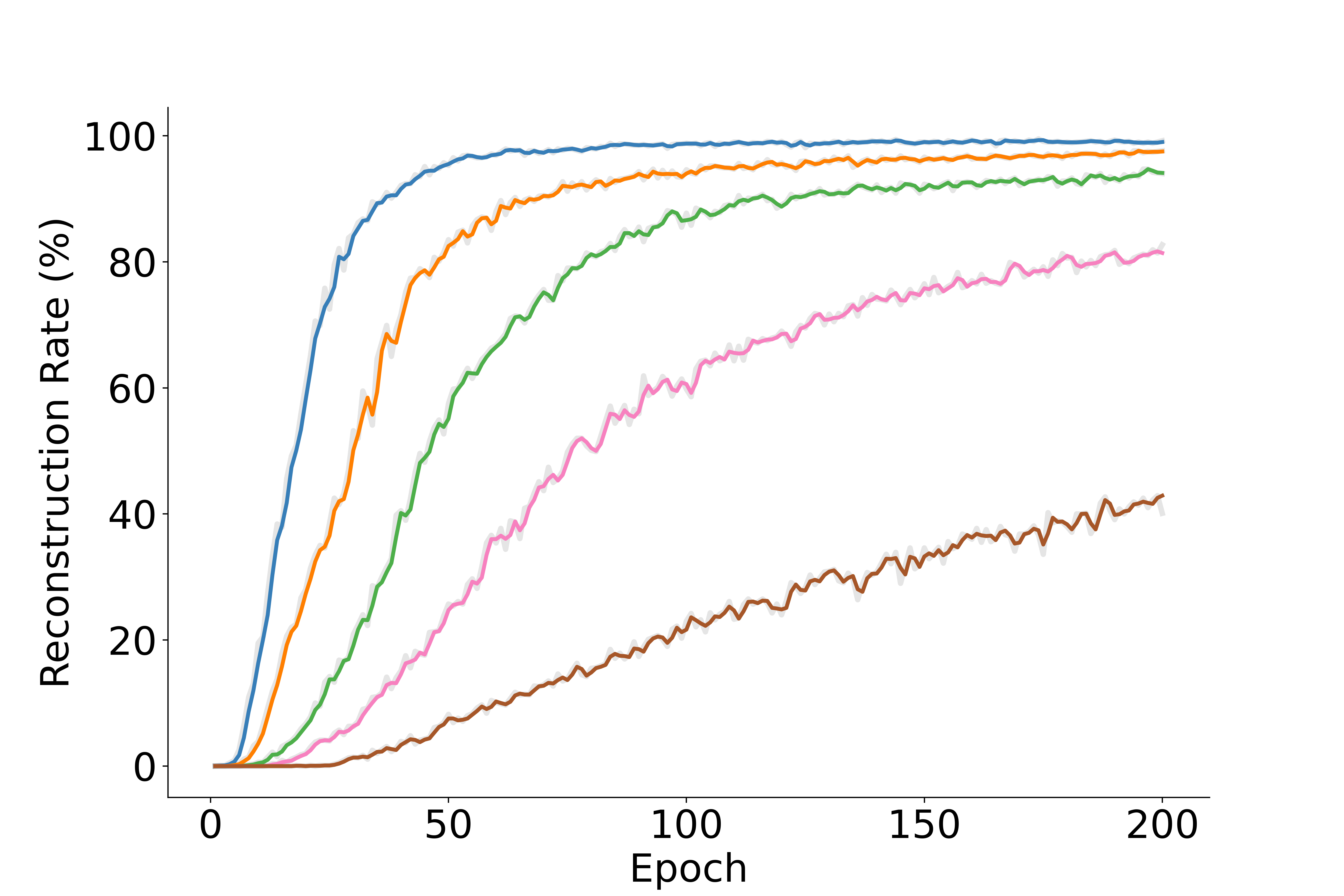} 
      \label{fig:KL_L_reconstruction_comparison}

    \end{subfigure}
    \begin{subfigure}{.475\textwidth}
      \caption{}
      \centering
      \includegraphics[width=\textwidth]{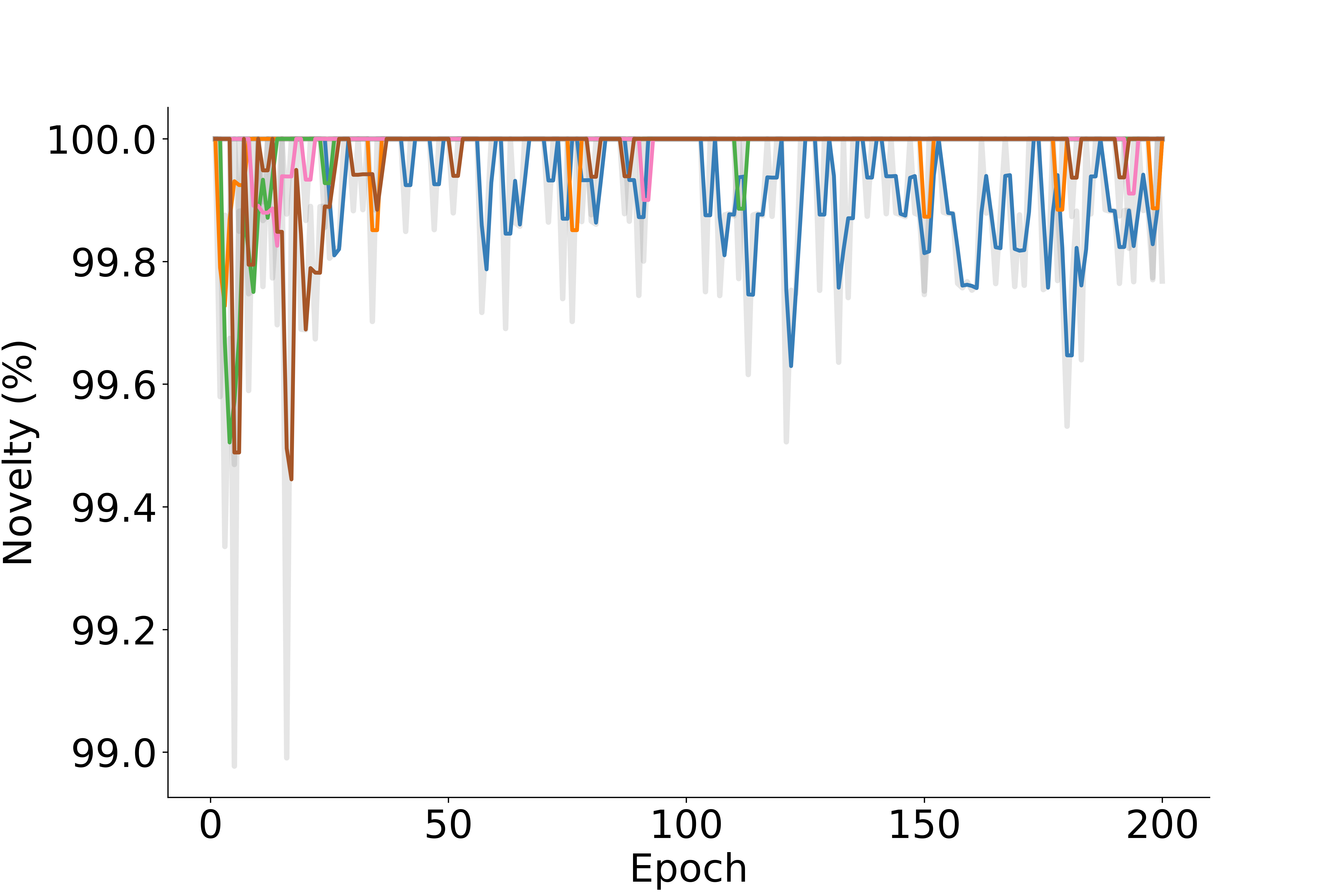} 
      \label{fig:KL_L_novelty_comparison}
    \end{subfigure}
    \begin{subfigure}{.475\textwidth}
      \caption{}
      \centering
      \includegraphics[width=\textwidth]{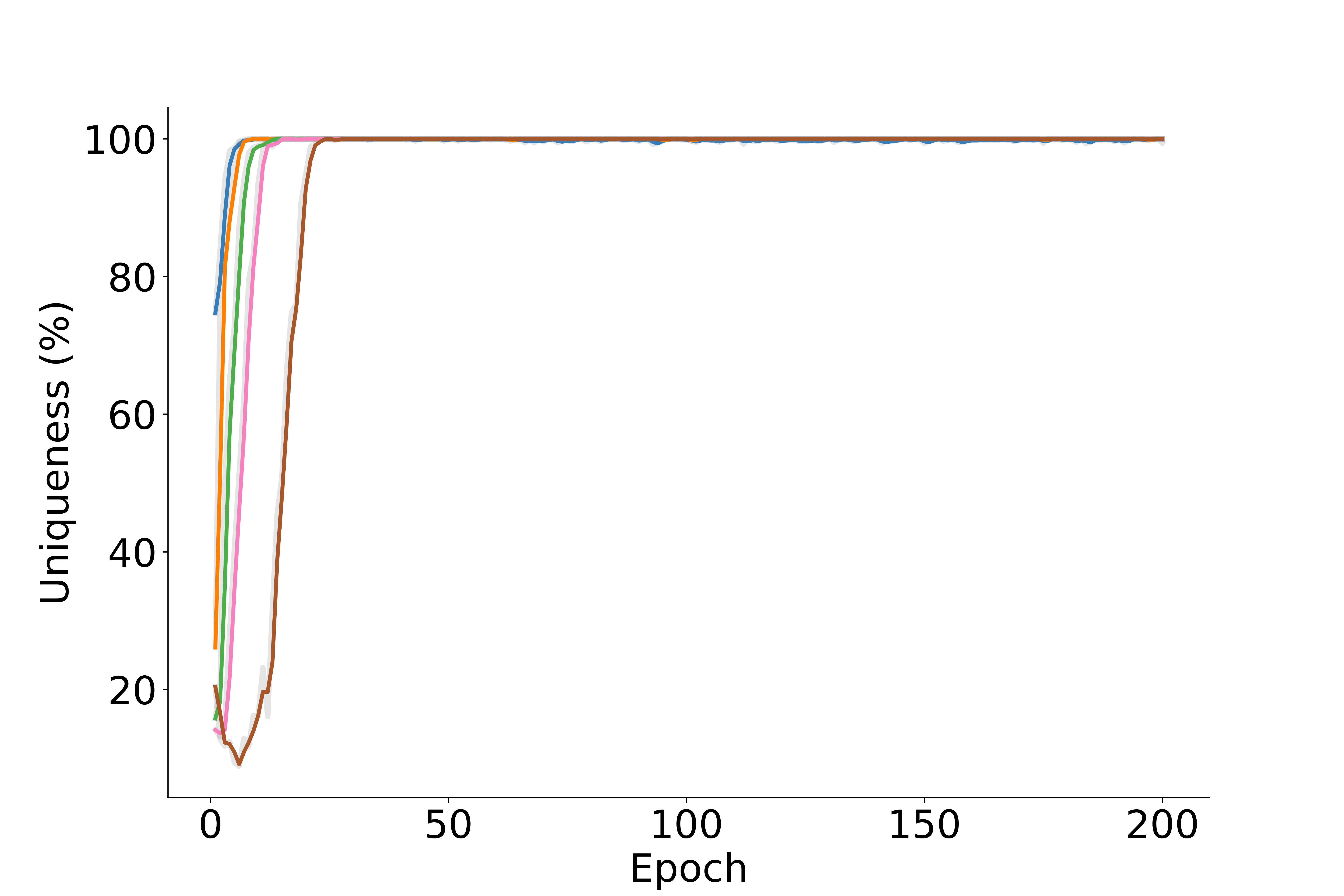} 
      \label{fig:KL_L_uniqueness_comparison}
    \end{subfigure}
    
    \hfill
    \begin{subfigure}{\textwidth}
      \centering
      \includegraphics[width=0.9\textwidth]{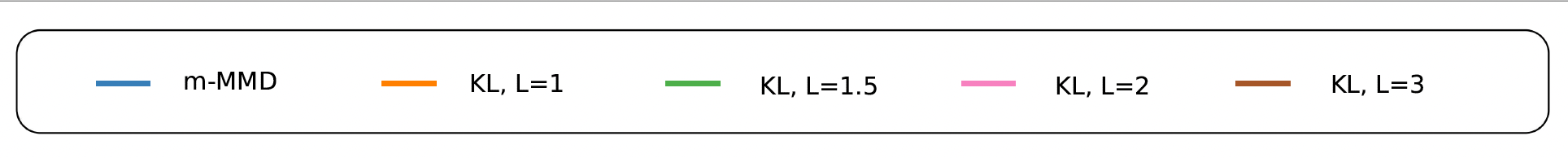} 
    \end{subfigure}
    \hfill

\caption{\textbf{Comparison of learning rates for models trained with different $\lambda$ values for KL divergence loss.} (a) Validity evaluated at each epoch. (b) Fraction of molecules properly reconstructed as a function of epochs. (c) Novelty evaluated at each epoch. (d) The uniqueness at each epoch. The models are trained with the loss $\mathcal{L}_{VAE}=\mathcal{L}_{CEL} + \textit{KL}(\lambda)$. The m-MMD model with the loss $\mathcal{L}_{CEL} + \textit{m-MMD}(\lambda=1)$.}

\label{fig:KL_Lambda_comparison}
\end{figure*}

In Figure~\ref{fig:KL_loss_comparison}, we initially set all $\lambda$ values to 1. However, this choice may not represent the optimal $\lambda$ value for KL models when conducting a fair comparison with m-MMD models. To explore the impact of varying $\lambda$ values on KL models, we conducted a similar analysis as depicted in Figure~\ref{fig:KL_Lambda_comparison}. The results indicate that higher $\lambda$ values tend to enhance validity at the expense of reduced reconstruction rates. Specifically, for the KL model, the optimal $\lambda$ value is found around 1.5, yielding an NUVR value of 0.51, whereas the model applying m-MMD,  at $\lambda$ = 1 , achieves a NUVR of 0.85. It's worth noting that although KL models incorporate an additional layer for variance prediction, the increase in parameters is minimal compared to the overall parameter count (1e-3 over the total parameter counts). Thus, this comparison appropriately underscores the advantage of the m-MMD loss in terms of the NUVR metric.

\section{Similarity
Exhaustion Search Procedures}\label{sec:SES}

The hyperparameters of SES include the beam size ($B$), the interval ($\delta_{s}$), the maximum increase in condition ($\Delta$), and the number of repetitions in Phase-two ($R$). In our implementation, the parameters were set as $B=15$, $\delta_s=0.1$, $\Delta=20$, and $R=4$.

\textbf{Condition Search}:
The Condition Search, the initial stage of SES, begins by assigning each molecule to be optimized, denoted as $m_{i}$, with its corresponding PLogP value as the initial condition $c_{i}$. The index $i$ represents the molecule's position. The latent vector $z_{i}$ is obtained through the encoding process.

During each step $s_{j}$, where $j$ starts from zero, a search is conducted for the vector $z_{i}$ with an adjusted condition $c_{i} + j \delta_{s}$. The concatenated vector of $z_{i}$ and the updated condition vector are then passed to the decoder. By utilizing beam search, a set of $B$ results is generated at each step. This process continues until the increment $j \delta_{s}$ reaches the maximum allowed value, $\Delta$. In total, $B + \frac{B\Delta}{\delta_{s}}$ candidates are produced for the molecule $m_{i}$ through this procedure. Subsequently, all candidates are filtered, retaining only those that exhibit a Tanimoto similarity within the range of 0.4 compared to the original molecule. The PLogP values of the remaining candidates are calculated and ranked. The optimization process is deemed successful if the highest PLogP value among the candidates for the $i^{\text{th}}$ molecule surpasses its original value. In such cases, the corresponding PLogP value and the SMILES representation of the candidate are recorded.

The purpose of condition search is to look for a set of candidates with similar encoder-estimated $z_{i}$ but with higher PLogP conditions. However, this procedure does not guarantee good samplings around all candidates. This means the decoded molecules would be dissimilar or even out of the similarity constraints from the encoded targets. In addition, despite the correct reconstruction, because these molecules represent the tail of the distribution of the PLogP conditions in the training data, they could have poor latent space definitions around them. This can cause a similar problem to reconstruction where better candidates within the constraint cannot be found due to a decrease in factors such as validity, uniqueness, and novelty. Therefore, a repositioning step is developed to ensure all molecules, especially for those $z_{i}$ that cannot be reconstructed correctly, can explore possibly better-starting points in the later search. 

\textbf{Repositioning}:
Repositioning is used to encourage sampling from regions farther away from the encoded latent vectors. To achieve this, sampling around the vector $z_{i}$ at the corresponding condition $c_{i}$ is performed. The sampling process involves adding a noise term $\epsilon$ drawn from the same Gaussian distribution employed during training. 

\begin{figure}[!h]
    \centering
    \includegraphics[width=0.5\textwidth]{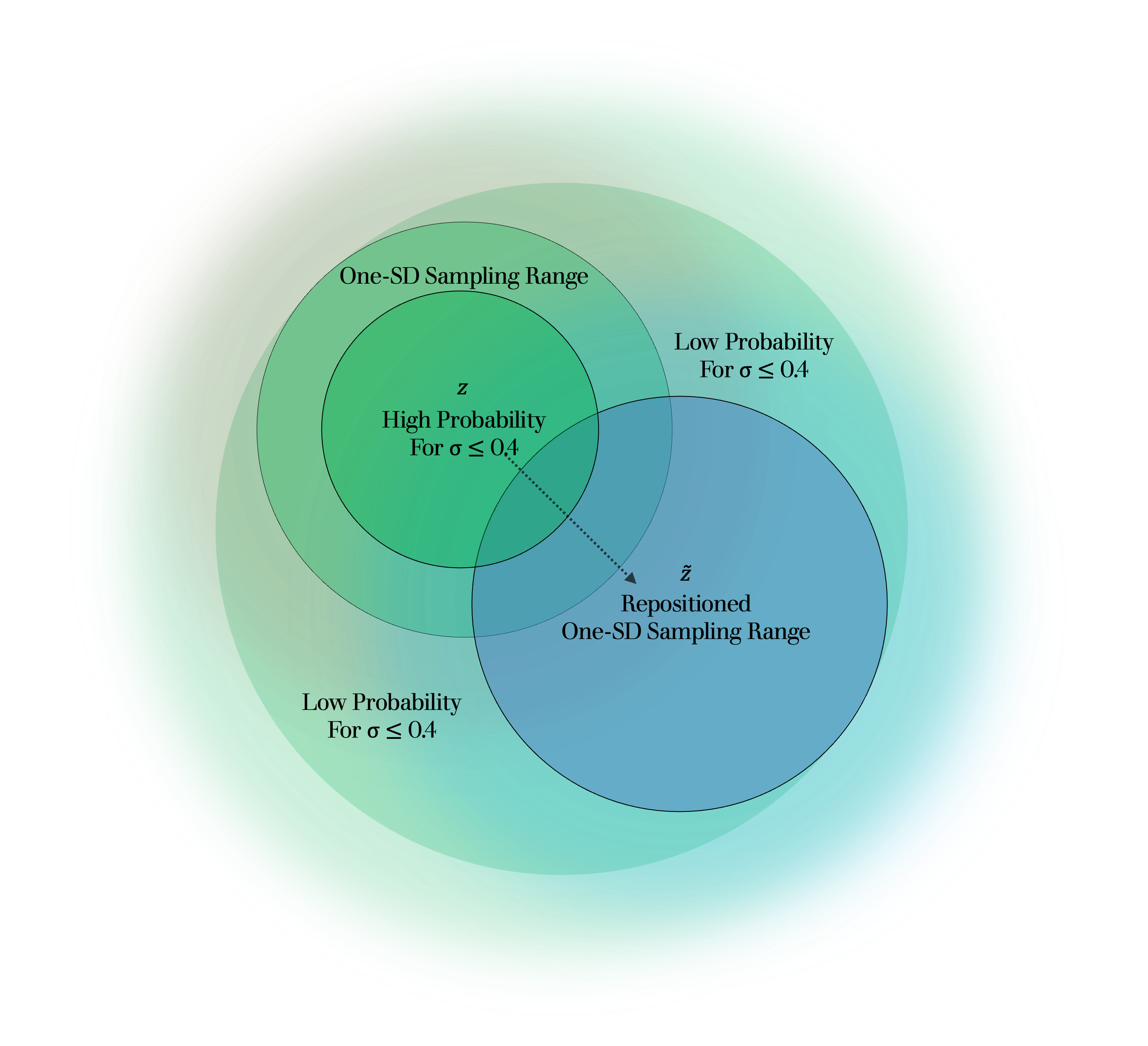}
    \caption{\textbf{Repositioning.} A $\tilde{z_{i}}$ is selected around $z_{i}$ if the generated molecule falls within the similarity threshold ($\sigma$) and exhibits an improvement in the optimized property. The subsequent search repetition is then conducted around $\tilde{z_{i}}$. Through repositioning, the search space expands for molecules that showed little improvements during the condition search.}
    \label{fig:Resampling}
\end{figure}

If the sampled vector $\tilde{z_{i}}$ yields a superior outcome compared to the previous search, it is recorded. Whenever a recorded $\tilde{z_{i}}$ exists, the subsequent sampling iteration starts from this repositioned vector. This repositioning step aims to expand exploration towards molecules that exhibit a greater separation from $z_{i}$, especially for vectors that display limited or no improvement during the condition search. This repositioning procedure is iterated 100 times to enhance the exploration process. A visual representation of this procedure can be seen in Figure ~\ref{fig:Resampling}.

\textbf{Phase Two}:
The preceding stages of the SES, namely the condition search and repositioning, yield two distinct sets of latent vectors. The first set comprises the original encoded $z$ vectors, while the second set consists of the repositioned $\tilde{z}$ vectors. In the second phase, the search process is performed in parallel using a combination of the condition search and repositioning approaches.

For each set, noise is added in a similar manner as during the repositioning stage. However, in this phase, every $c_{i}$ is adjusted by $c_{i} + j \delta_{s}$, following the same procedure as the condition search.

By applying the filtering and selection criteria identical to those used in the condition search, new molecules with the highest PLogP values are recorded for both sets. The phase two process is repeated $R$ times. After completing the $R$ repetitions, for each molecule candidate, the superior result between the two sets is chosen.

\section{Impact of the $\lambda$ Parameter}
\label{section:lambda}

The reason for increasing $\lambda$ is similar to that of increasing $\beta$ in the case for $\beta$-VAE by Richards et al. Both $\lambda$ in KAE and $\beta$ in  $\beta$-VAE encourage the model to learn more efficient latent representations and to construct smoother latent space. However, since KAE has different architecture and loss objectives from VAE, the aforementioned regularisation does not lead to the same result in terms of the NUVR metric when both $\lambda$ and $\beta$ are set to one for KAE and $\beta$-VAE.

For the best model using m-MMD in Figure ~\ref{fig:KL_loss_comparison}, all validities are lower than 90\%. This can be improved by increasing the $\lambda$ value for the m-MMD term as shown in Table~\ref{table:lambda}. The models in Table~\ref{table:lambda} were first trained with $\lambda=1$ for 85 epochs then with higher values for an additional 1 epoch. $\delta$ values were set to $-\lambda$ throughout the training process to exclude any effects from $\mathcal{L}_{WCEL}$ in the comparison.

Increasing the $\lambda$ value tightens the placement of latent vectors together in the Gaussian distribution penalized by the m-MMD loss. This is reflected by the increase in the probability of sampling valid molecules when the latent vectors are drawn from the same distribution. However, as the latent vectors become closer, it becomes more challenging for the decoder to differentiate them, resulting in a decrease in reconstruction. The decrease in uniqueness and novelty with increased $\lambda$ values can be attributed to the decoder more frequently identifying different molecules with overlapping latent representations as the same ones.

The overall effects of $\lambda$ are shown by the NUVR metrics. Table~\ref{table:lambda} shows the trend of NUV and NUVR as $\lambda$ is adjusted. It is observed that validity peaks with larger $\lambda$ 
and the model trained with $\lambda=24.5$ has the highest NUV. However, the reconstruction rate decreases significantly with increasing $\lambda$ values.

\begin{table*}[!ht]
\caption{\textbf{Model performance with varying $\lambda$.} The result shows sampling 1k latent vectors by continued training of the model from the same checkpoint (85 epochs) with the loss function being $\mathcal{L}(\lambda=1, \delta=-1)$, but then followed by an additional epoch with different $\lambda$ values (loss functions are then $\mathcal{L}(\lambda=\lambda, \delta=-\lambda)$).}
  \centering
  
  \begin{tabular}{ccccccc}
    \toprule
    \cmidrule(r){1-2}
    $\lambda$   & Validity  & Novelty     & Uniqueness & NUV & Reconstruction & NUVR\\
    \midrule
    1.0 &  0.782  & 1.000 & 0.995 & 0.778  & \textbf{0.988} & 0.769   \\
    2.0 &  0.802  & 1.000 & 1.000 & 0.802  & 0.978 & 0.784   \\
    5.0 &  0.849  & 1.000 & 1.000 & 0.849  & 0.933 & \textbf{0.792}   \\
    10.0 & 0.847  & 0.999 & 0.999 & 0.845  & 0.792 & 0.669   \\
    15.0 & 0.913  & 0.998 & 1.000 & 0.911  & 0.527 & 0.480   \\
    20.0 & 0.929  & 1.000 & 1.000 & 0.929  & 0.246 & 0.229   \\
    24.5 & 0.961  & 0.999 & 0.998 & \textbf{0.958}  & 0.060 & 0.057  \\
    25.0 & 0.940  & 0.998 & 1.000 & 0.938  & 0.043 & 0.040   \\
    25.5 & 0.943  & 1.000 & 1.000 & 0.943  & 0.039 & 0.037   \\
    26.0 & 0.965  & 0.998 & 0.999 & 0.962  & 0.029 & 0.028   \\
    27.5 & 0.962  & 0.996 & 0.999 & 0.957  & 0.010 & 0.010  \\
    30.0 & \textbf{0.970}  & 1.000 & 0.987 & 0.957  & 0.000 & 0.000  \\
    \bottomrule
  \end{tabular}

  \label{table:lambda}
\end{table*}

\begin{figure*}
    \centering
    \captionsetup[subfigure]{position=top, labelfont=bf, textfont=normalfont, singlelinecheck=off, justification=raggedright}
    \begin{subfigure}{0.45\textwidth}
      \caption{}
      \centering
      \includegraphics[width=0.9\textwidth]{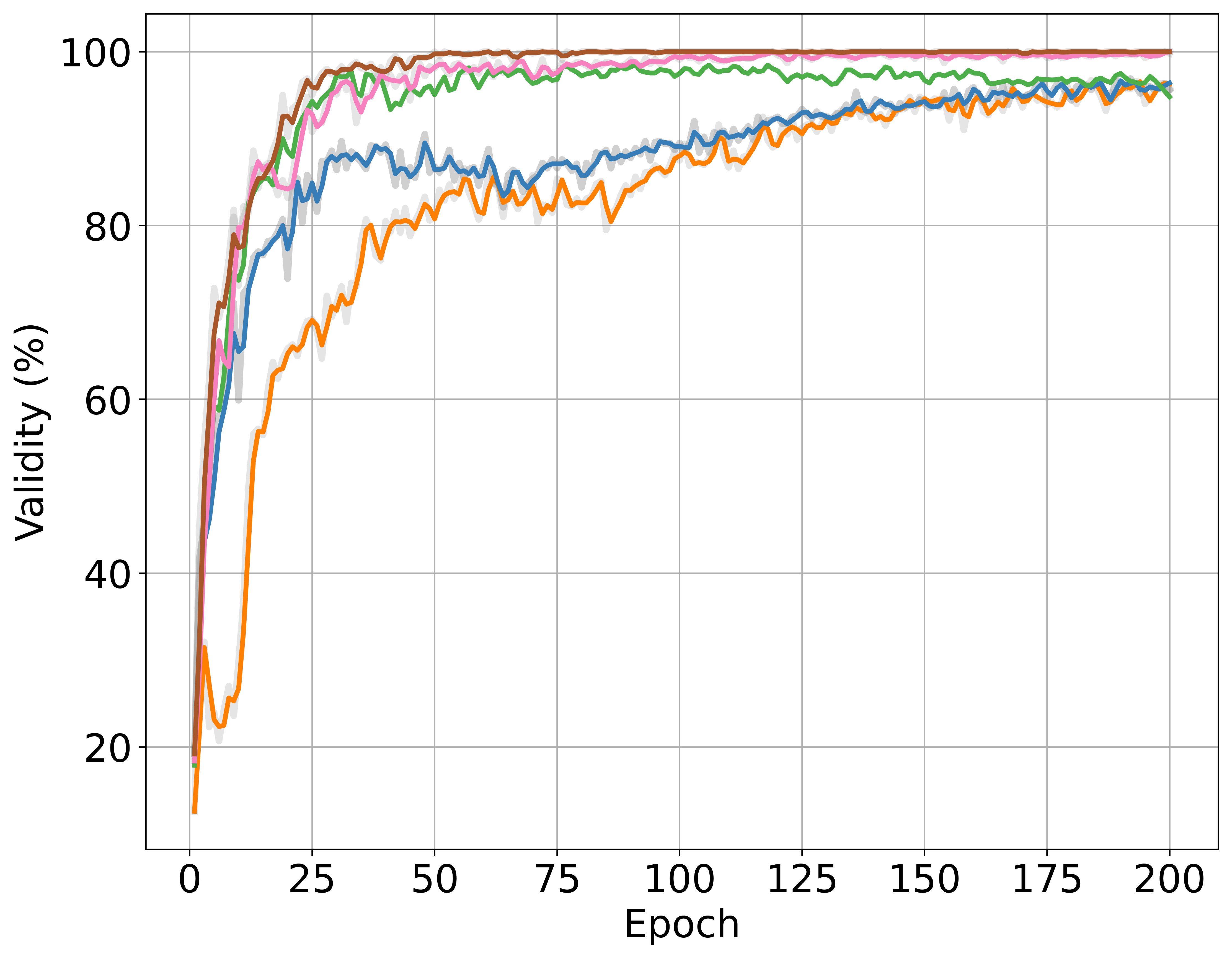} 
      \label{fig:lambda_validity_comparison}
    \end{subfigure}
    \quad
    \begin{subfigure}{0.45\textwidth}
      \caption{}
      \centering
      \includegraphics[width=0.9\textwidth]{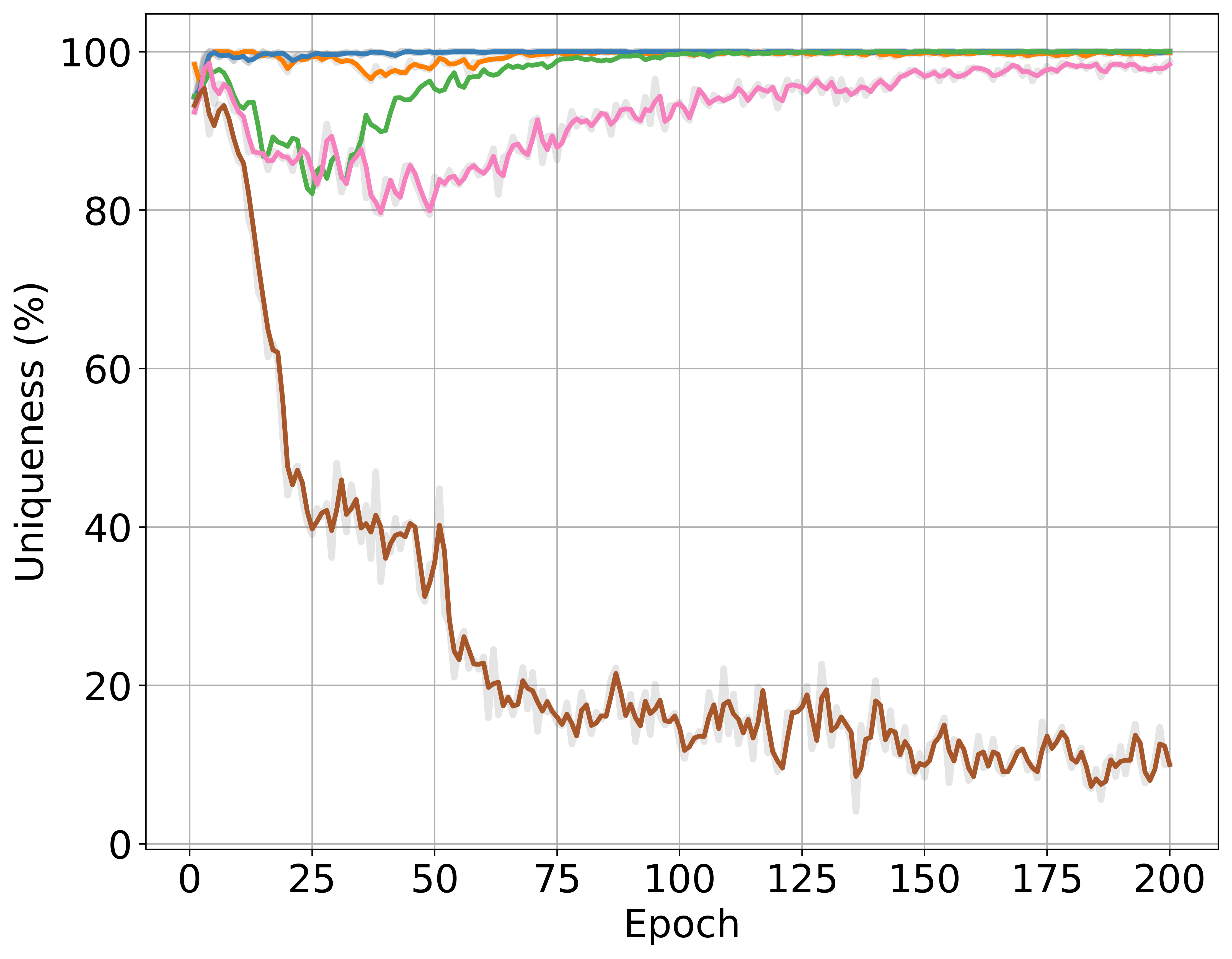}  
      \label{fig:lambda_uniqueness_comparison}
    \end{subfigure}

    \begin{subfigure}{0.45\textwidth}
      \caption{}
      \centering
      \includegraphics[width=0.9\textwidth]{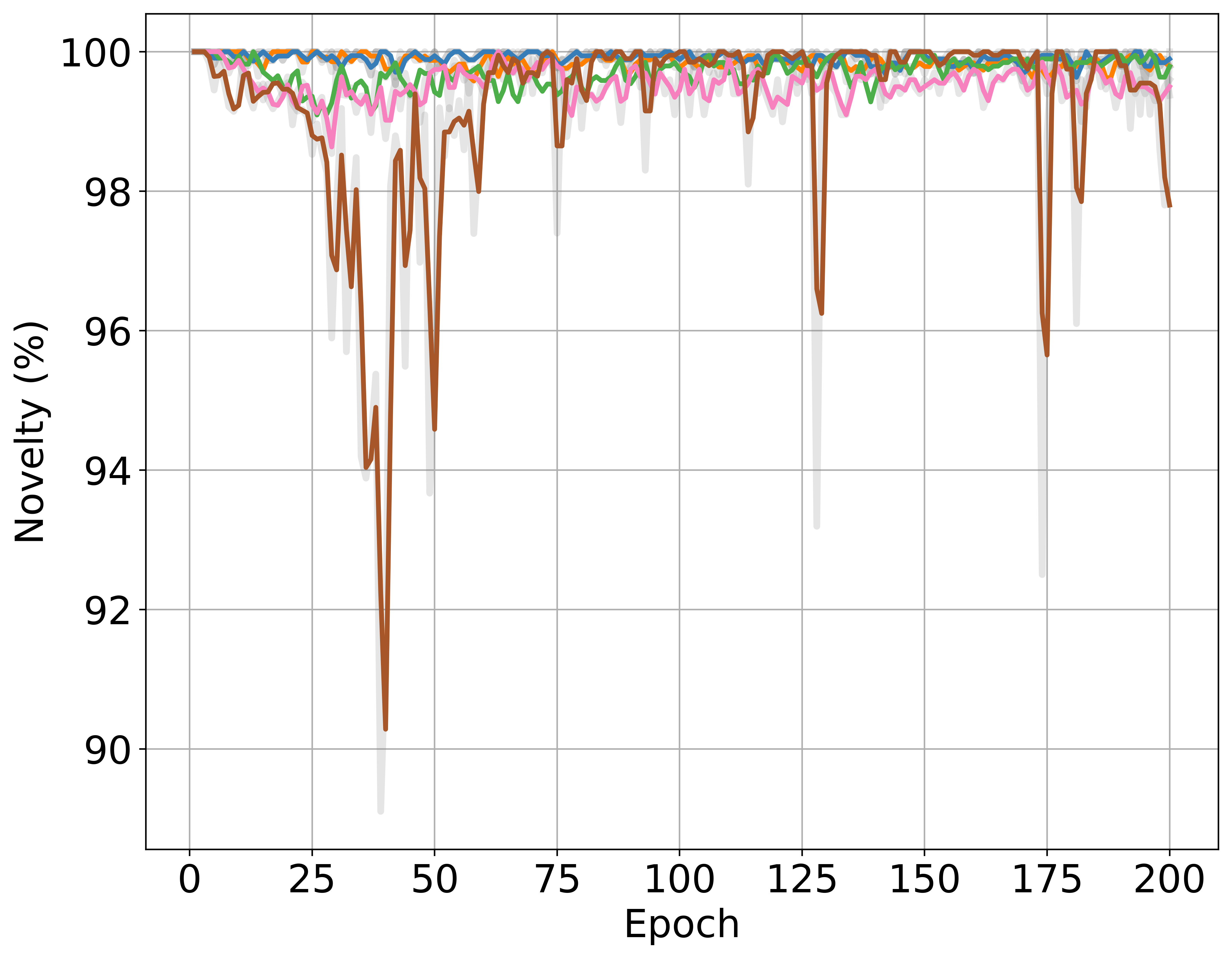}  
      \label{fig:lambda_novelty_comparison}
    \end{subfigure}
    \quad
    \begin{subfigure}{0.45\textwidth}
      \caption{}
      \centering
      \includegraphics[width=0.9\textwidth]{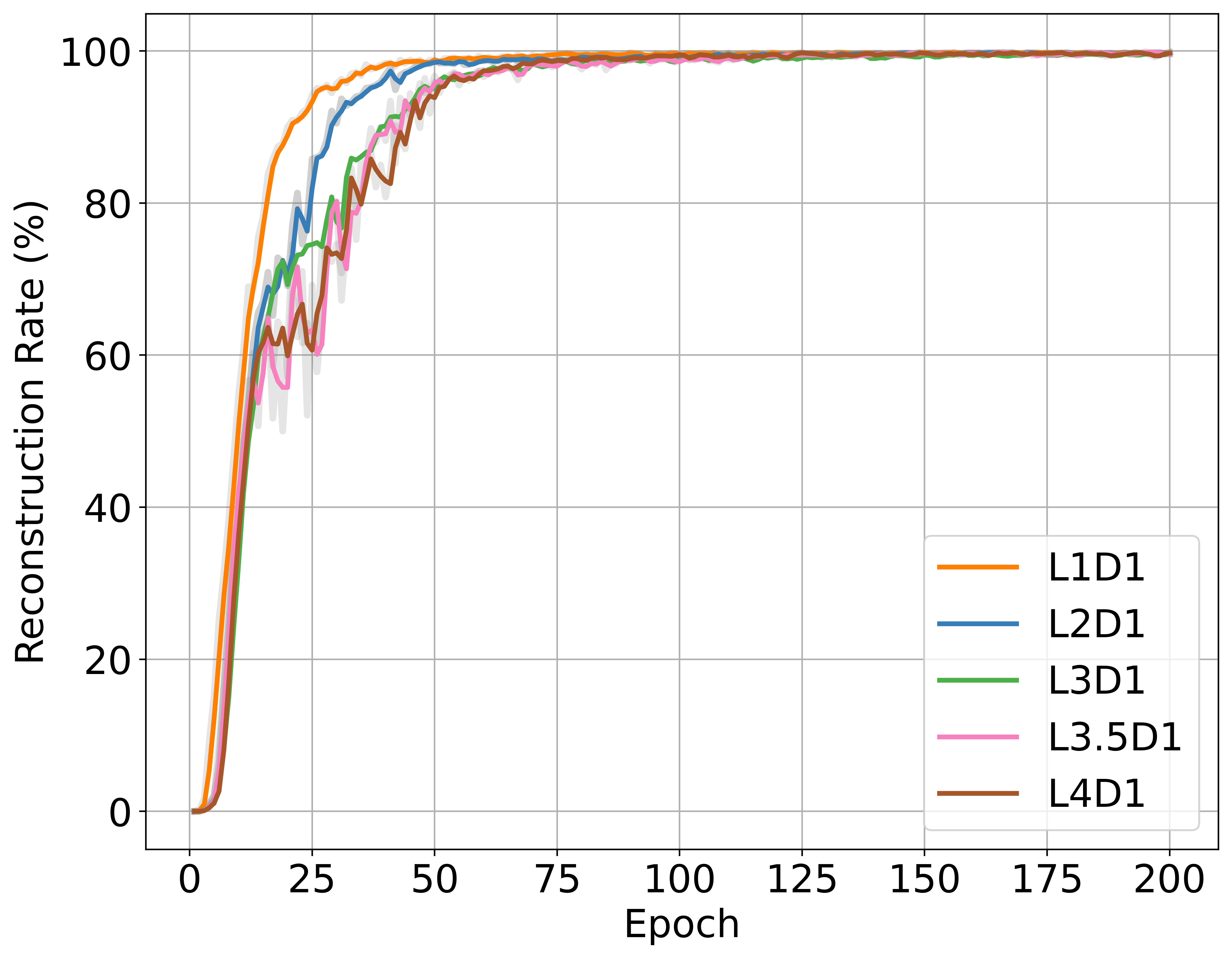} 
      \label{fig:lambda_reconstruction_comparison}
    \end{subfigure}

\caption{\textbf{The performance comparison of models trained with different $\lambda$ values, while keeping $\delta=1$, using the m-MMD loss.} (a) Validity evaluated at every epoch. (b) Uniqueness evaluated at every epoch. (c) Novelty evaluated at every epoch. (d) Reconstruction rate evaluated at every epoch. The evaluation metrics include validity, uniqueness, and novelty, which are computed at the end of each epoch based on 1000 randomly generated molecules from each model. Additionally, the reconstruction rate is calculated using 1000 molecules from the validation set. In the legend, the notation LxDy represents a model trained with $\lambda=x$ and $\delta=y$. For instance, the model labeled L3D1 corresponds to $\mathcal{L}(\lambda=3, \delta=1)$.}

\label{fig:lambda_comparison}
\end{figure*}

We seek to find a solution that can increase validity while maintaining other metrics at the same level. Therefore controlling the model via $\mathcal{L}_{WCEL}$ was the key to this problem. Model performance with a range of $\delta$ values were compared in \ref{fig:delta_comparison} and $\delta=1$ was chosen in $\mathcal{L}_{WCEL}$; We then compared different $\lambda$ values with $\delta$ fixed to 1. In Figure ~\ref{fig:lambda_comparison}, models are trained with the loss function $\mathcal{L}(\lambda, \delta=1)$ for 200 epochs. It can be observed in Figure ~\ref{fig:lambda_validity_comparison} that higher $\lambda$ values lead to better final validity. However, uniqueness in Figure ~\ref{fig:lambda_uniqueness_comparison} breaks down for the case of when $\lambda = 4$ while novelty and reconstruction rates converge to around 100\% in (Figure ~\ref{fig:lambda_novelty_comparison} and Figure ~\ref{fig:lambda_reconstruction_comparison}). Therefore, the $\lambda = 3.5$ model is trained for additional 200 epochs (total 400 epochs) to give final performance metrics in Table~\ref{table:NUVR_comparison}.

\section{Latent Space And Model Performance}\label{sec:Latent_Space_And_Performance}

\begin{figure*}
    \centering
    \captionsetup[subfigure]{position=top, labelfont=bf, textfont=normalfont, singlelinecheck=off, justification=raggedright}
    
    \begin{subfigure}{0.45\textwidth}
      \caption{}
      \centering
      \includegraphics[width=0.9\textwidth]{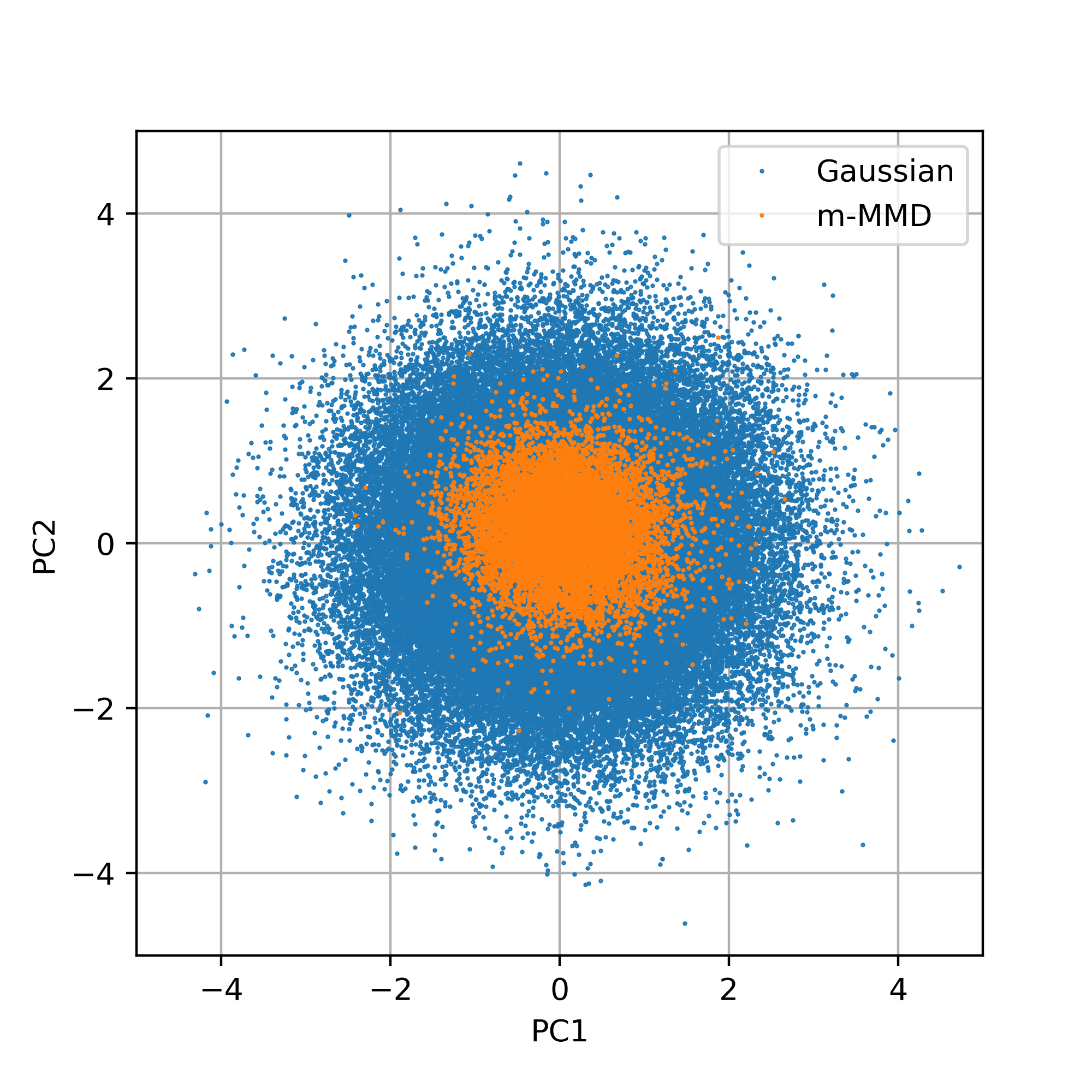} 
      \label{fig:m-MMD_Latent}
    \end{subfigure}
    \quad
    \begin{subfigure}{0.45\textwidth}
      \caption{}
      \centering
      \includegraphics[width=0.9\textwidth]{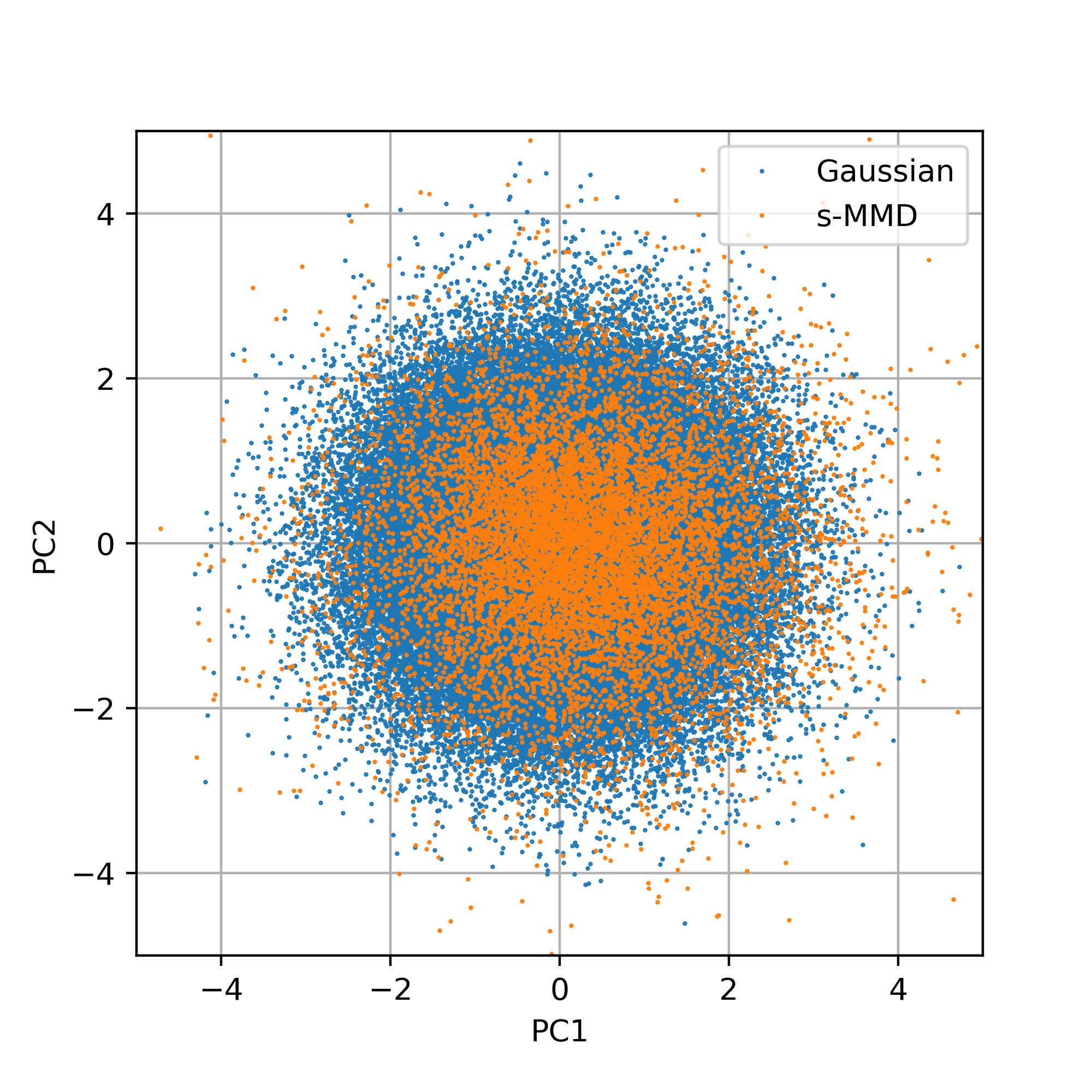}
      \label{fig:s-MMD_Latent}
    \end{subfigure}

    \begin{subfigure}{0.45\textwidth}
      \caption{}
      \centering
      \includegraphics[width=0.9\textwidth]{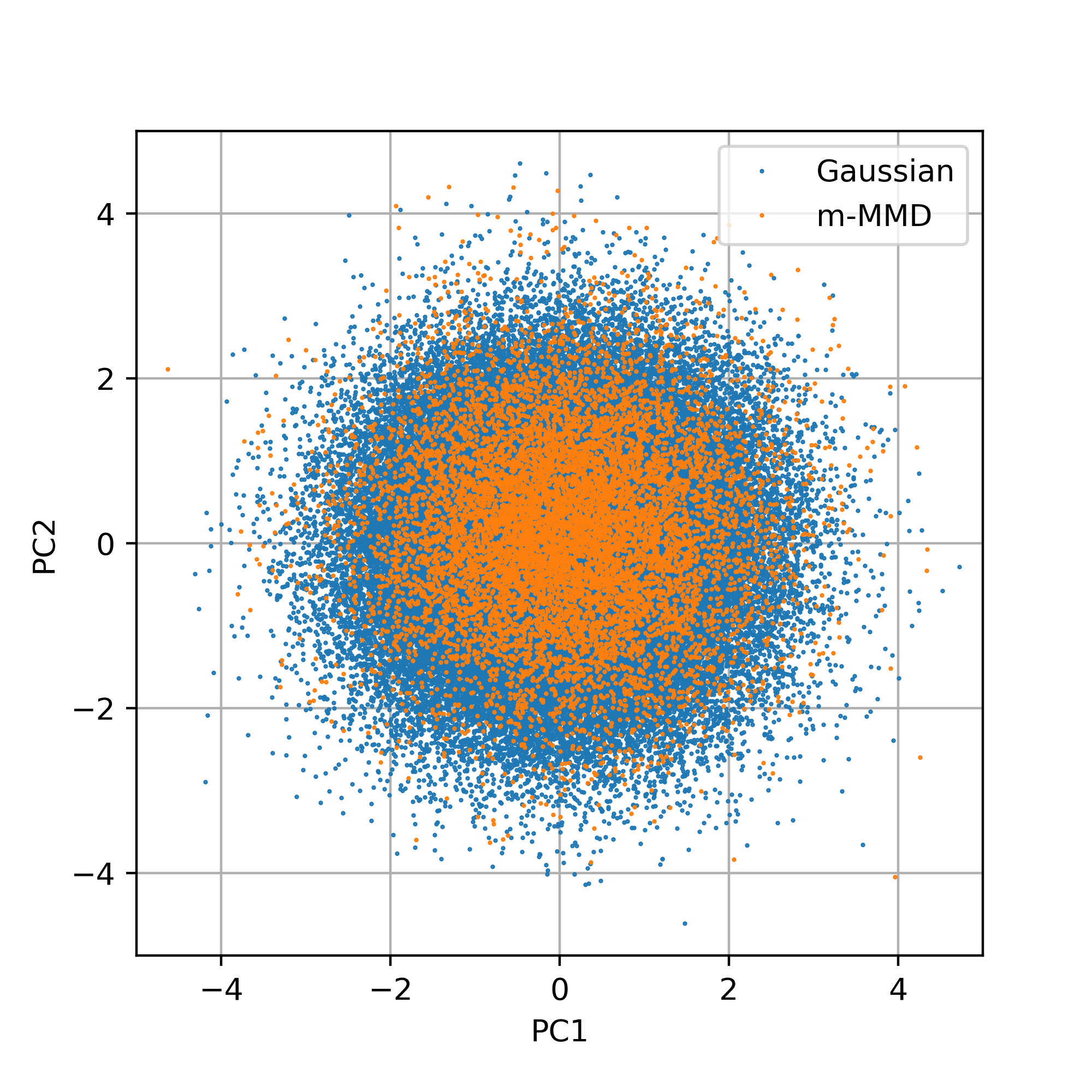} 
      \label{fig:m-MMD_Latent_with_noise}
    \end{subfigure}
    \quad
    \begin{subfigure}{0.45\textwidth}
      \caption{}
      \centering
      \includegraphics[width=0.9\textwidth]{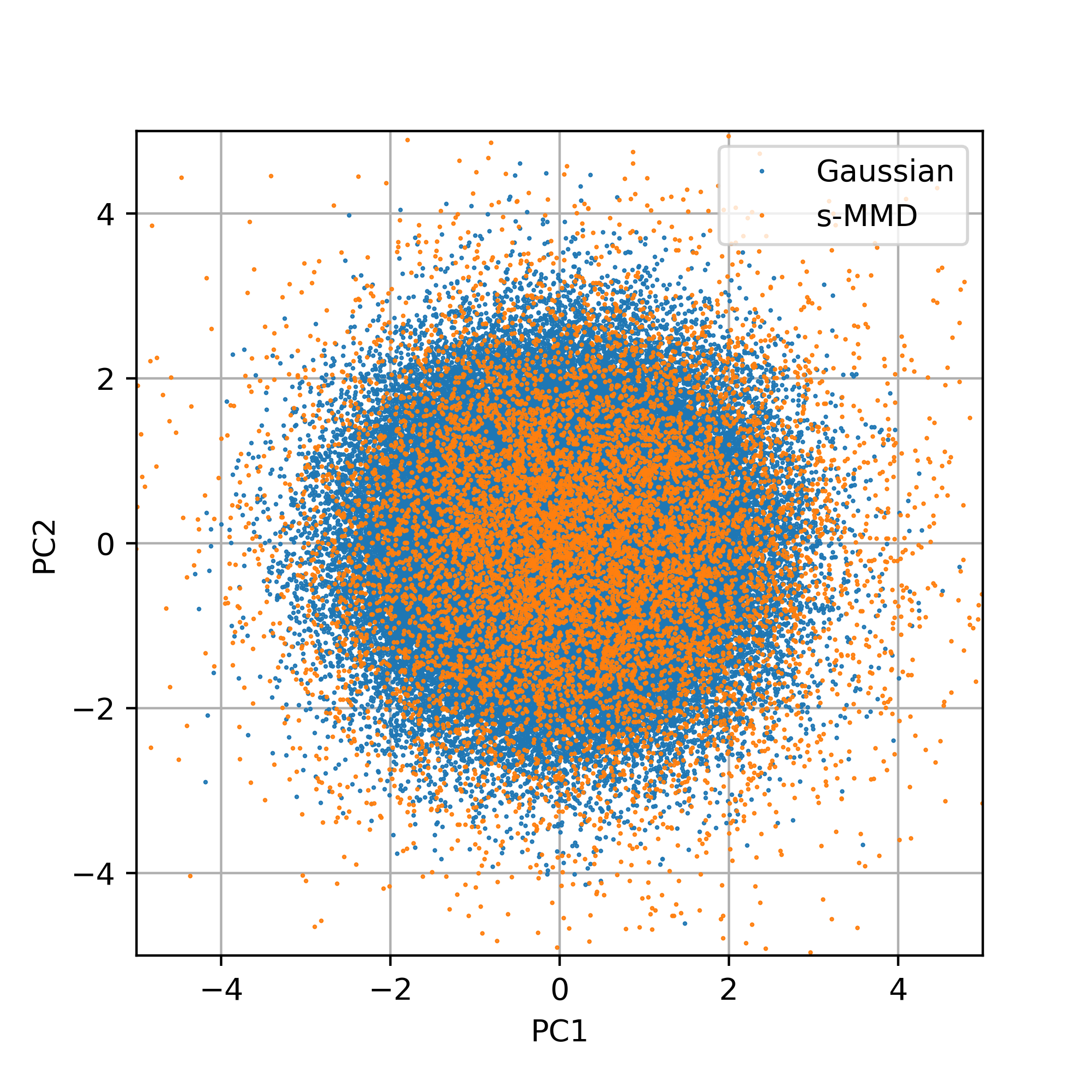}
      \label{fig:s-MMD_Latent_with_noise}
    \end{subfigure}

\caption{\textbf{Latent vectors obtained by passing 10k ZINC250K molecules to the encoder and transforming under the same principal components extracted from the standard Gaussian distribution.} (a) m-MMD results showing all latent vectors well-incorporated in the Gaussian. (b) s-MMD loss makes the latent vectors more scattered relative to the Gaussian, making it less likely to obtain valid output by sampling from the Gaussian. (c) and (d) show the actual vectors passed into the decoder in the training process with latent noise added.}
\label{fig:Latent_Comparison}
\end{figure*}

In m-MMD, with the RBF-kernel function, removing the $\vec{\mu_{x}}^{T} \vec{\mu_{x}}$ term is believed to be helpful since as it allows the distributions of individual molecules to be closer together. This makes the sampling region have fewer places where the decoder cannot infer valid molecules. 
A demonstration and a comparison with the latent spaces of s-MMD and m-MMD is presented in Figure ~\ref{fig:Latent_Comparison}.

It can be seen from Figure ~\ref{fig:reconstruction_comparison} that, with or without noise, the models trained with s-MMD have a faster-converging reconstruction rate than the models trained with m-MMD. This is because the extra $\mathcal{K}(\vec{x},\vec{x})$ term in s-MMD promotes the separation of the latent representations of the data points such that the decoder can easily differentiate the representations. However, since the latent vectors that represent valid molecules are far from each other, the validity is significantly lower for the models trained with s-MMD.

We consider increasing $\lambda$ as an approach to optimize the model performance in N, U, and V, reducing the uninterpreted regions while still making individual molecules distinct from each other.

\begin{figure*}
    \centering
    \begin{subfigure}{0.475\textwidth}
      \centering
      \includegraphics[width=\textwidth]{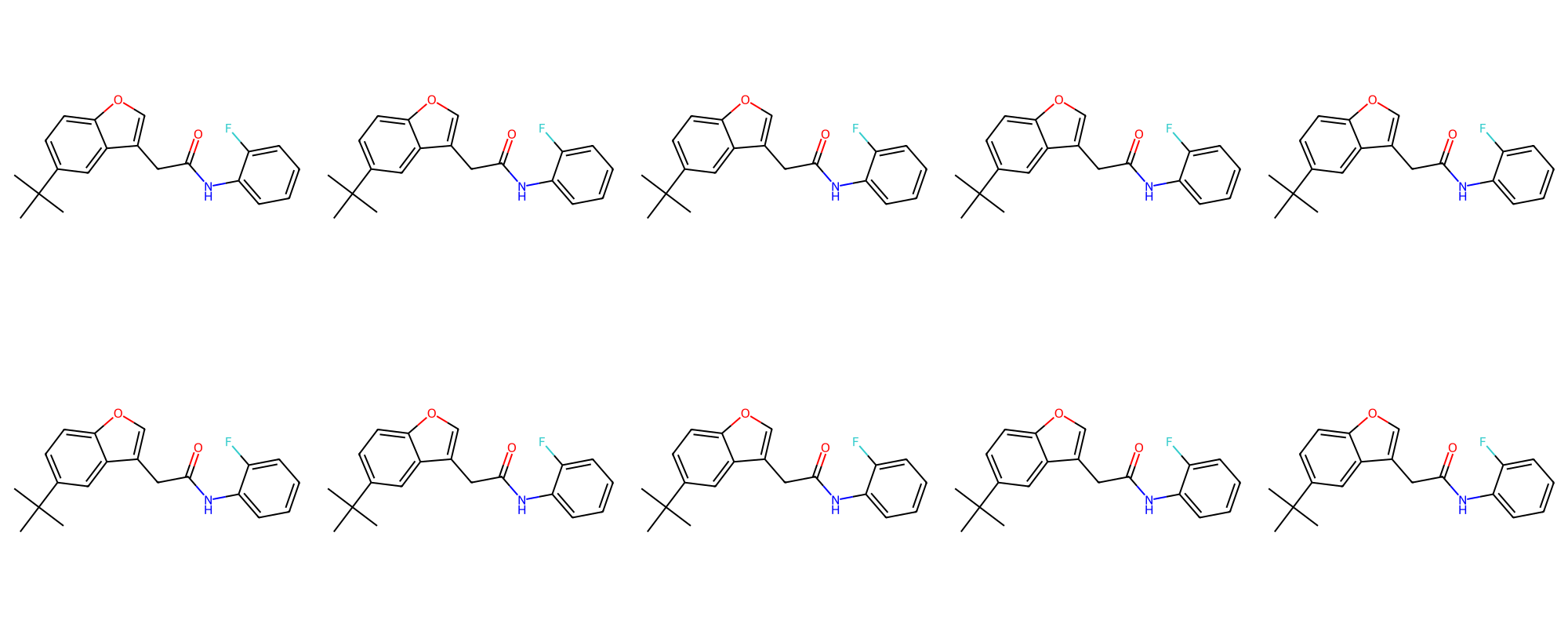} 
      \caption{Molecules found with a beam size of one}
      \label{fig:B1_01STD}
    \end{subfigure}
    \hfill
    \begin{subfigure}{0.475\textwidth}
      \centering
      \includegraphics[width=\textwidth]{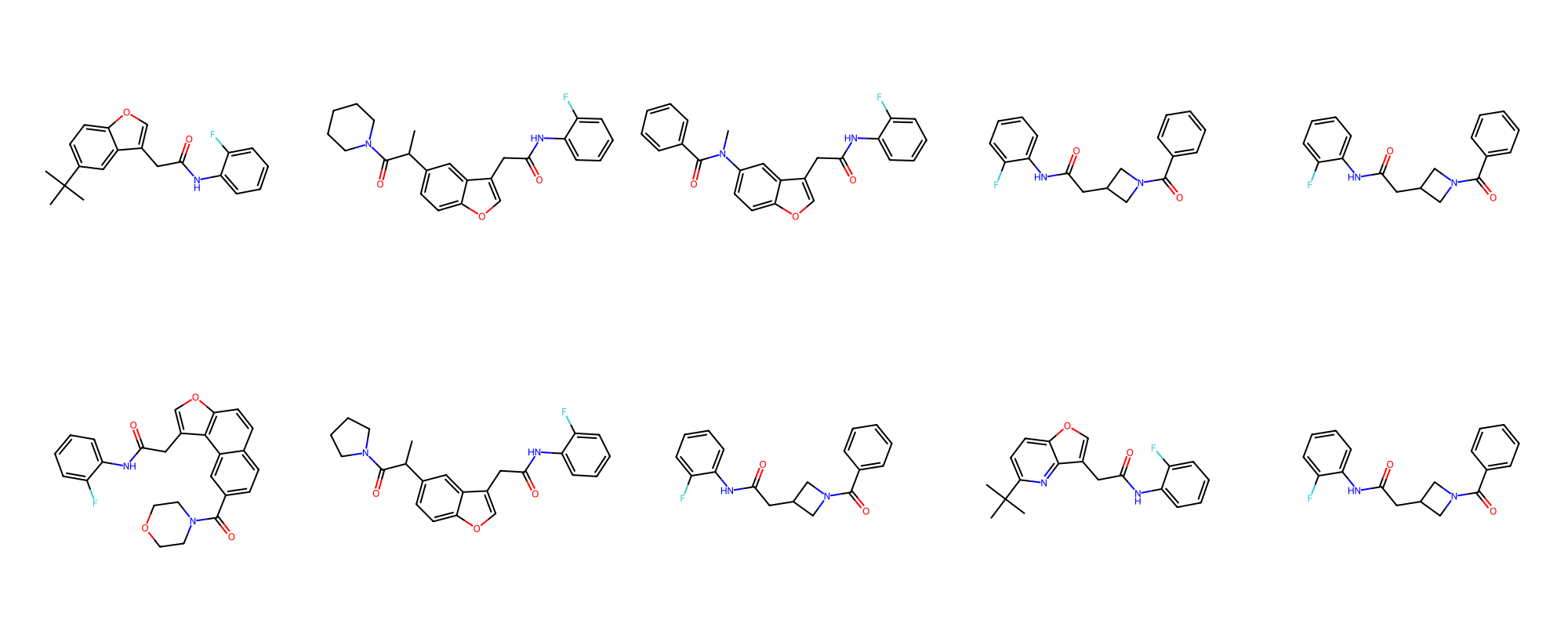}  
      \caption{Molecules found with a beam size of two}
      \label{fig:B2_01STD}
    \end{subfigure}
    \hfill
    
\caption{\textbf{Molecules obtained by sampling from the 0.1-SD Gaussian distribution centered around a specific latent vector, while varying the beam size.} Figure (a) illustrates the molecules found using a beam size of one, where only the original encoded molecule is identified despite the added noise. In contrast, Figure (b) showcases the molecules discovered when a beam size of two is employed, revealing seven distinct molecules out of the ten samples.}
\label{fig:Beam_1vs2}
\end{figure*}

\section{Enhancing Generation Performance through Beam Search}
To further improve the performance of our model, we employ beam search, a popular decoding technique in sequence generation tasks. Beam search involves selecting a single output from a set of $B$ potential candidates, based on specific criteria outlined in the decoding method (see Section~\ref{sec:decoding_method}).

\begin{table*}[!h]
\caption{\textbf{Model performance with varying beam sizes.} This table shows the model's generation performance with various beam sizes by sampling 10k latent vectors each time. One output is selected out of the interpretations given by all beam search results for each latent vector. The result with a beam size of one is an equivalent measurement to other methods that do not use beam search and grammar checks.}
  \centering
  \begin{tabular}{ccccc}
    \toprule
    \cmidrule(r){1-2}
    Beam Size     & Novelty     & Uniqueness & Validity & NUV \\
    \midrule
    
    1 & 0.996  & 0.974 & 0.998 & 0.968 \\
    \hline
    2 & 1.000  & 0.996 & 1.000 & 0.996  \\
    \hline
    3 & 1.000  & 0.998 & 1.000 & 0.998   \\
    \hline
    4 & 1.000  & 1.000 & 1.000 & 1.000  \\
    \hline
    5 & 1.000  & 1.000 & 1.000 & 1.000  \\
    \bottomrule
  \end{tabular}

  \label{table:Beam}
\end{table*}

Similar to previous studies that have employed checking methods to enhance model performance, we propose a new approach with beam search as a post-generation evaluation step. During the molecule generation process, we consider one of the outputs obtained from the beam search results. For instance, with a beam size of two, two possible interpretations are generated for the same latent vector. We iterate through the $B$ results, starting with the top-ranked interpretation. If any of the generated SMILES strings are novel, unique, and valid, the evaluation process is stopped and the corresponding SMILES string is retained. Priority is given to retaining valid molecules over those that are unique and novel. By checking and selecting from all the beam-searched outputs, we increase the likelihood of finding SMILES strings that meet the criteria of novelty, uniqueness, and validity. For a latent vector, if all its beam search results fail to meet the validity criterion, the top-one result is returned. We believe the beam search can help differentiate two latent vectors that are similar by providing alternative interpretations per vector.


In the case where the beam size is equal to one, the method is identical to greedy search which takes only the next-step candidate with maximum probability; We compare the results done at different beam sizes. 10k vectors are sampled for each listed beam size. The result of the beam size of one is used as the control group.

Table~\ref{table:Beam} presents the model's generation performance across different beam sizes, as measured by various metrics. The metrics assessed include novelty, uniqueness, validity, and the combined metric NUV. The results clearly indicate that as the beam size increases, the model's performance improves consistently. Notably, when the beam size exceeds three, the performance reaches a plateau, achieving the highest possible value of 1.0 for the NUV metric.

To further highlight the capabilities of beam search, we conduct additional experiments where we sample from a small distribution around a specific latent vector Fig~\label{fig:Beam_1vs2}. We start by selecting a molecule from the training set and encoding it into its corresponding latent vector. Next, we generate 10 noise vectors by sampling from a Gaussian distribution with a standard deviation one-tenth of that used during training (0.1-SD). These noisy latent vectors are then decoded using the beam search approach. The results obtained from beam search demonstrate the ability to find diverse candidates that are similar to the molecule being sampled. Notably, when a beam size of two is employed, six additional candidates are discovered compared to the case where beam search is not utilized (i.e., beam size of one).

\section{$\sigma$ Comparison}\label{sec:sigma_comparison}

\begin{figure*}
    \centering
    \captionsetup[subfigure]{position=top, labelfont=bf, textfont=normalfont, singlelinecheck=off, justification=raggedright}
    \begin{subfigure}{0.45\textwidth}
      \caption{}
      \centering
      \includegraphics[width=0.9\textwidth]{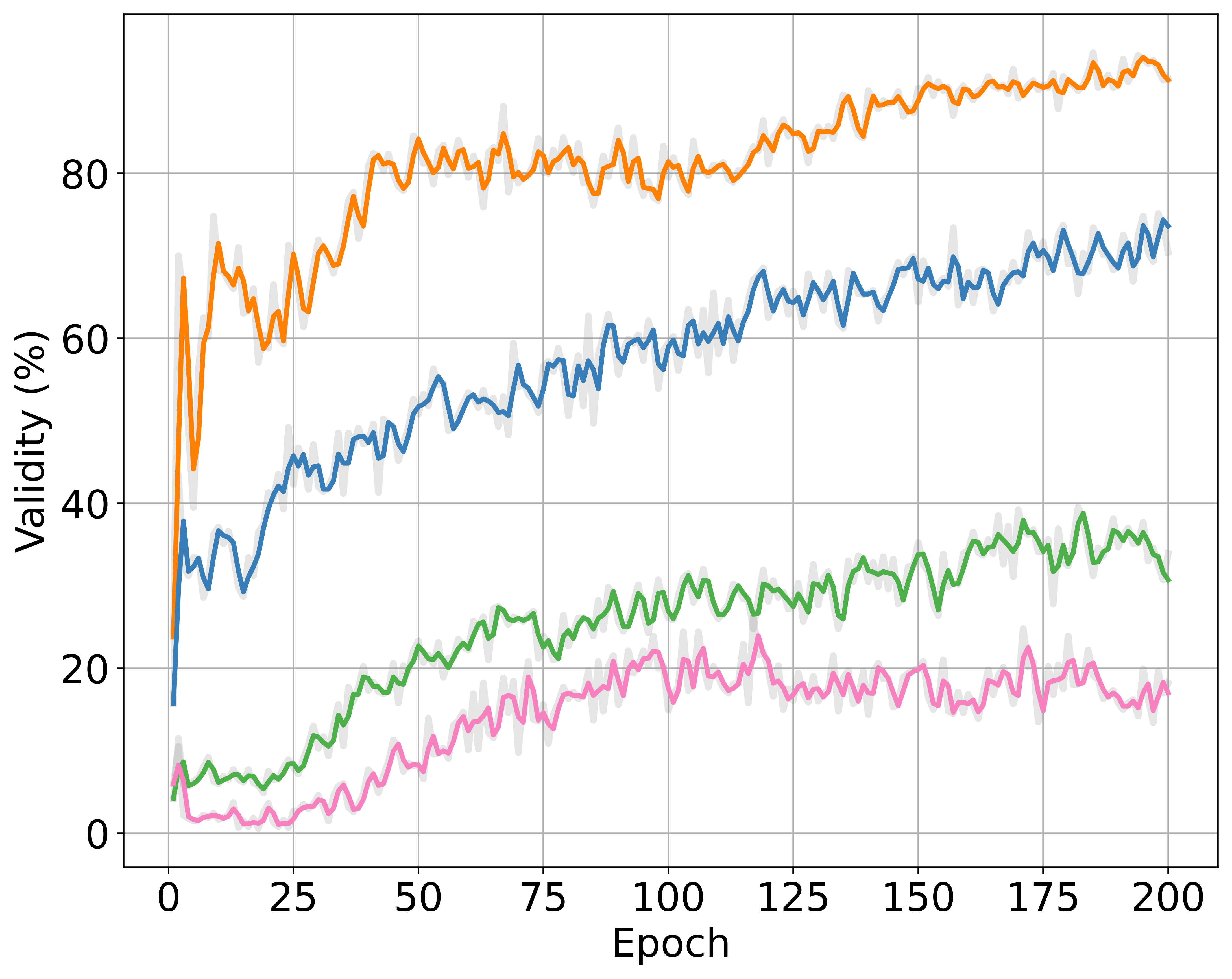} 
      \label{fig:sigma_mMMD_validity_comparison}
    \end{subfigure}
    \quad
    \begin{subfigure}{0.45\textwidth}
      \caption{}
      \centering
      \includegraphics[width=0.9\textwidth]{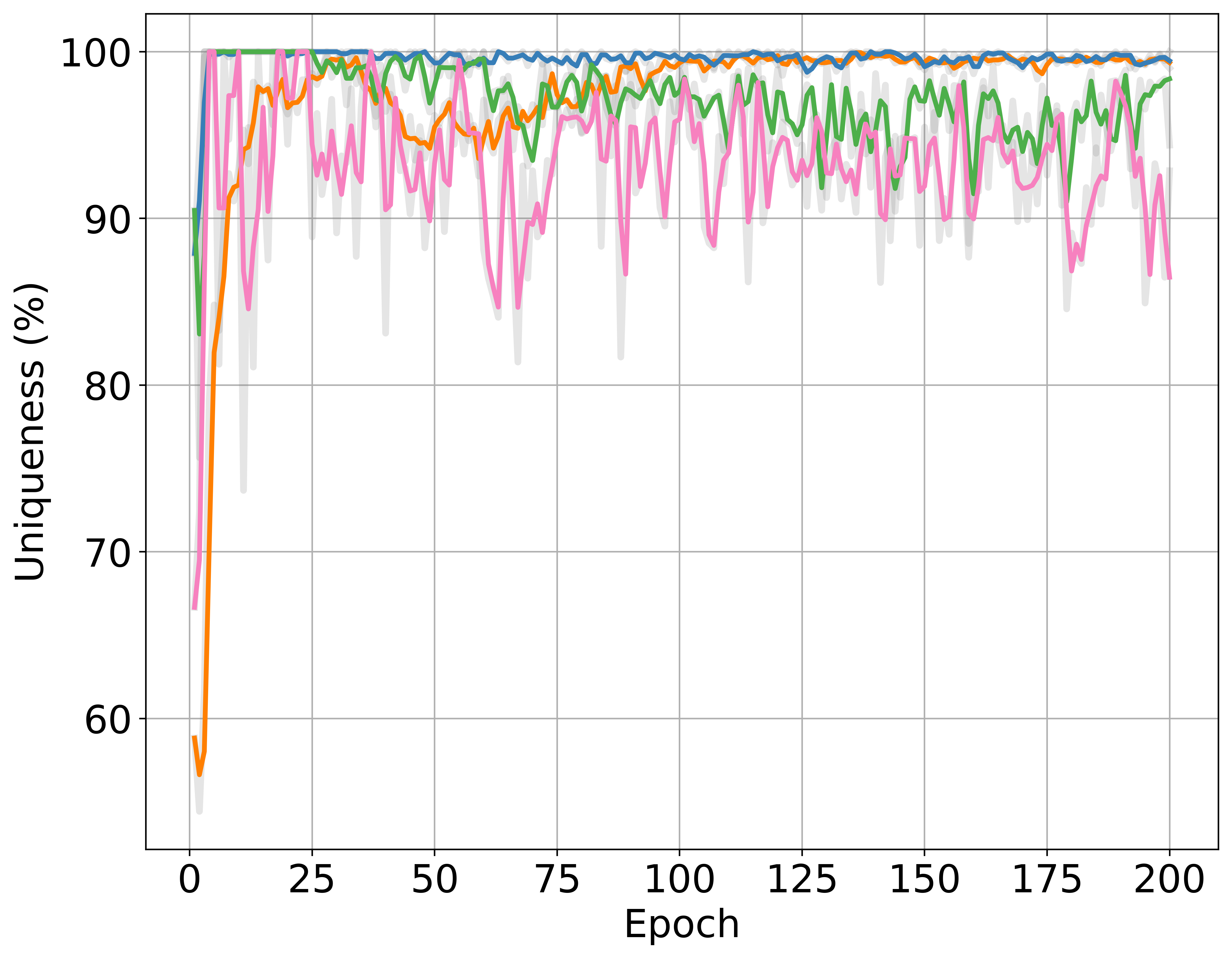}  
      \label{fig:sigma_mMMD_uniqueness_comparison}
    \end{subfigure}

    \begin{subfigure}{0.45\textwidth}
      \caption{}
      \centering
      \includegraphics[width=0.9\textwidth]{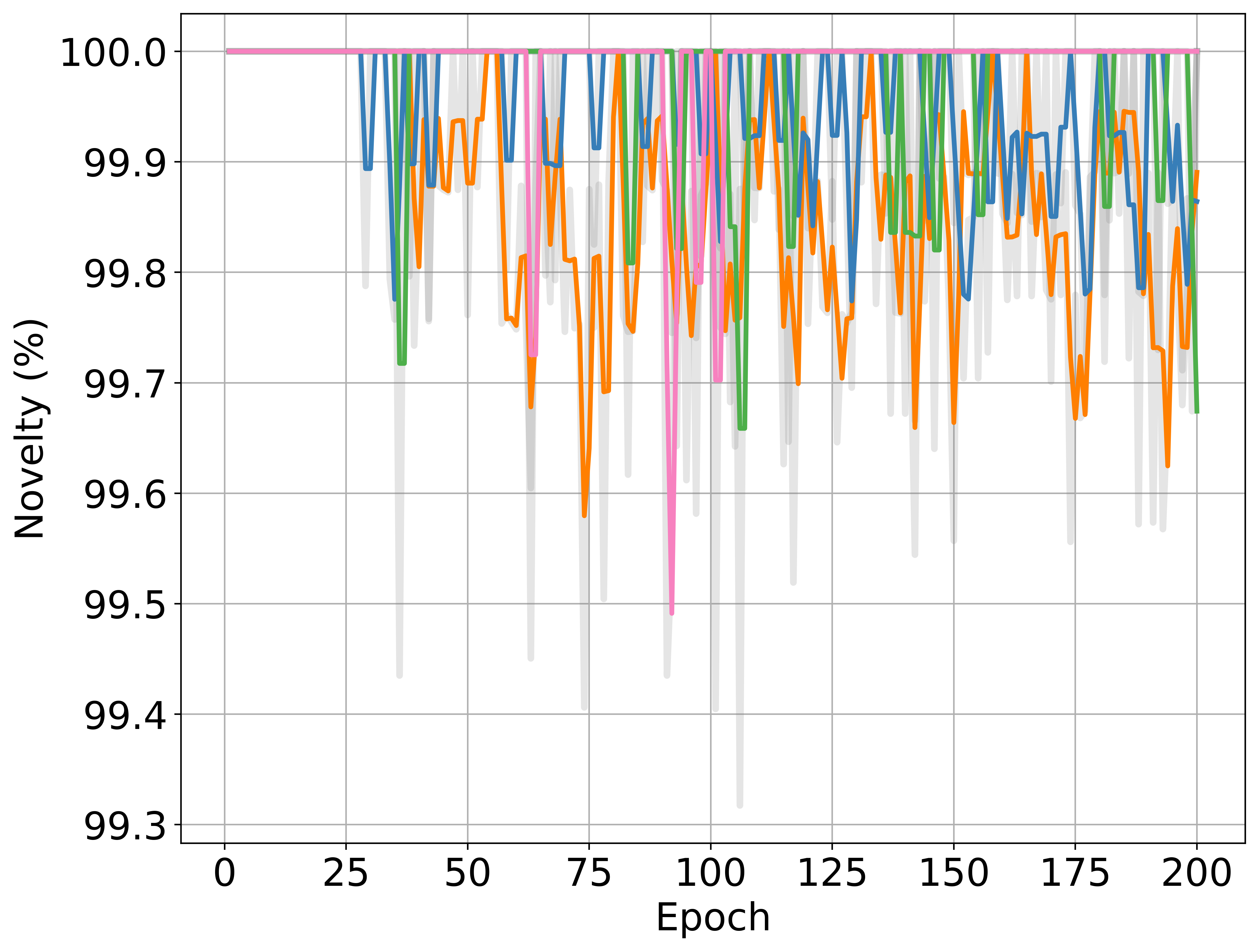}  
      \label{fig:sigma_mMMD_novelty_comparison}
    \end{subfigure}
    \quad
    \begin{subfigure}{0.45\textwidth}
      \caption{}
      \centering
      \includegraphics[width=0.9\textwidth]{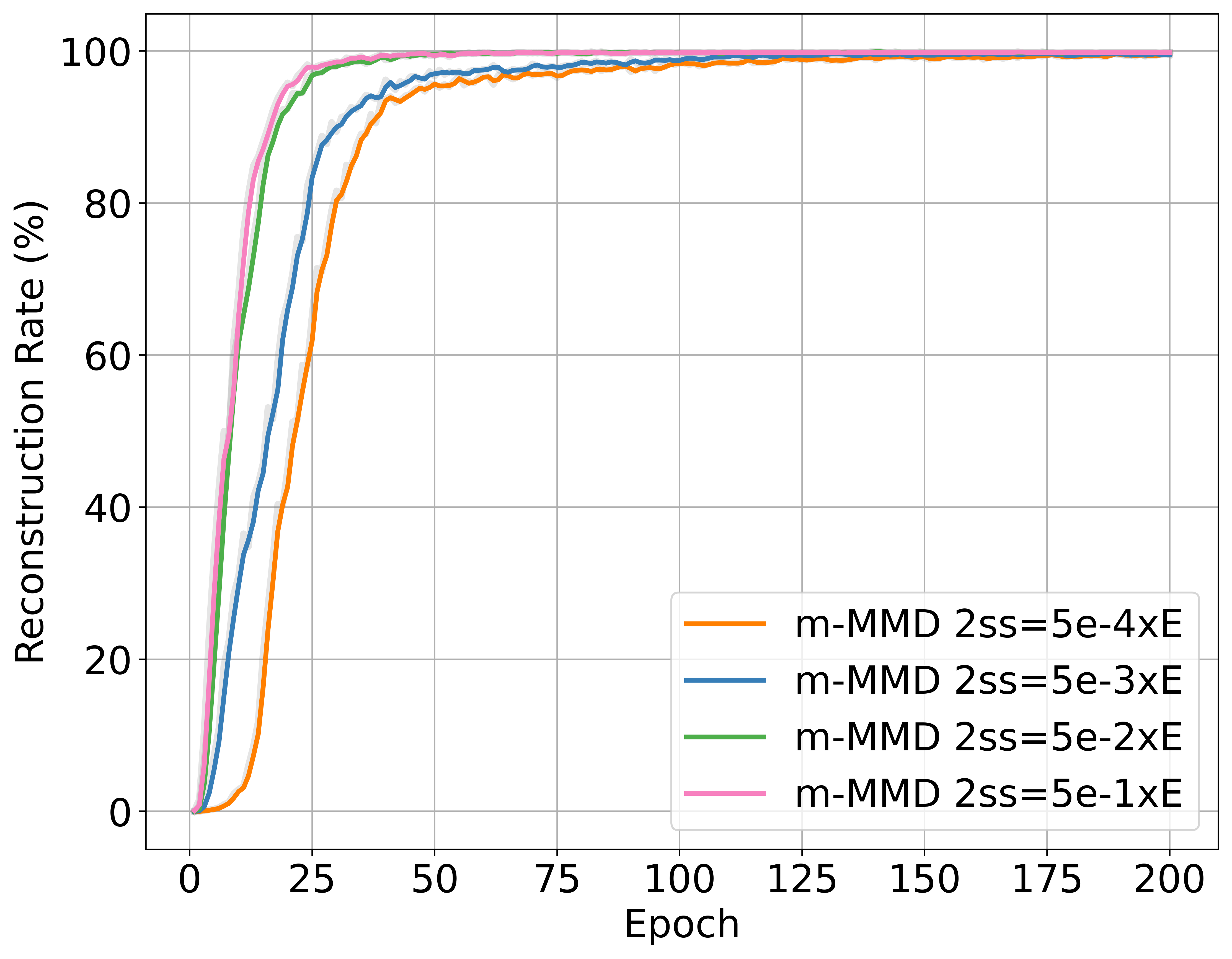} 
      \label{fig:sigma_mMMD_reconstruction_comparison}
    \end{subfigure}

\caption{\textbf{Performance comparison of the models trained with different sigma values using modified MMD loss.} (a) Validity evaluated at every epoch. (b) Uniqueness evaluated at every epoch. (c) Novelty evaluated at every epoch. (d) Reconstruction rate evaluated at every epoch. Note that 2ss (2 sigma squared) in the legend represents the value used for $2\sigma^2$ in Eq.~\ref{eq:Kernel} and $E$ is the embedding dimension. Validity, uniqueness, and novelty are calculated at the end of each epochs using 1000 randomly generated molecules from each of the models. And the reconstruction rate is calculated using 1000 molecules from the validation set.}
\label{fig:sigma_mMMD_comparison}
\end{figure*}

\begin{figure*}
    \centering
    \captionsetup[subfigure]{position=top, labelfont=bf, textfont=normalfont, singlelinecheck=off, justification=raggedright}
    \begin{subfigure}{0.45\textwidth}
      \caption{}
      \centering
      \includegraphics[width=0.9\textwidth]{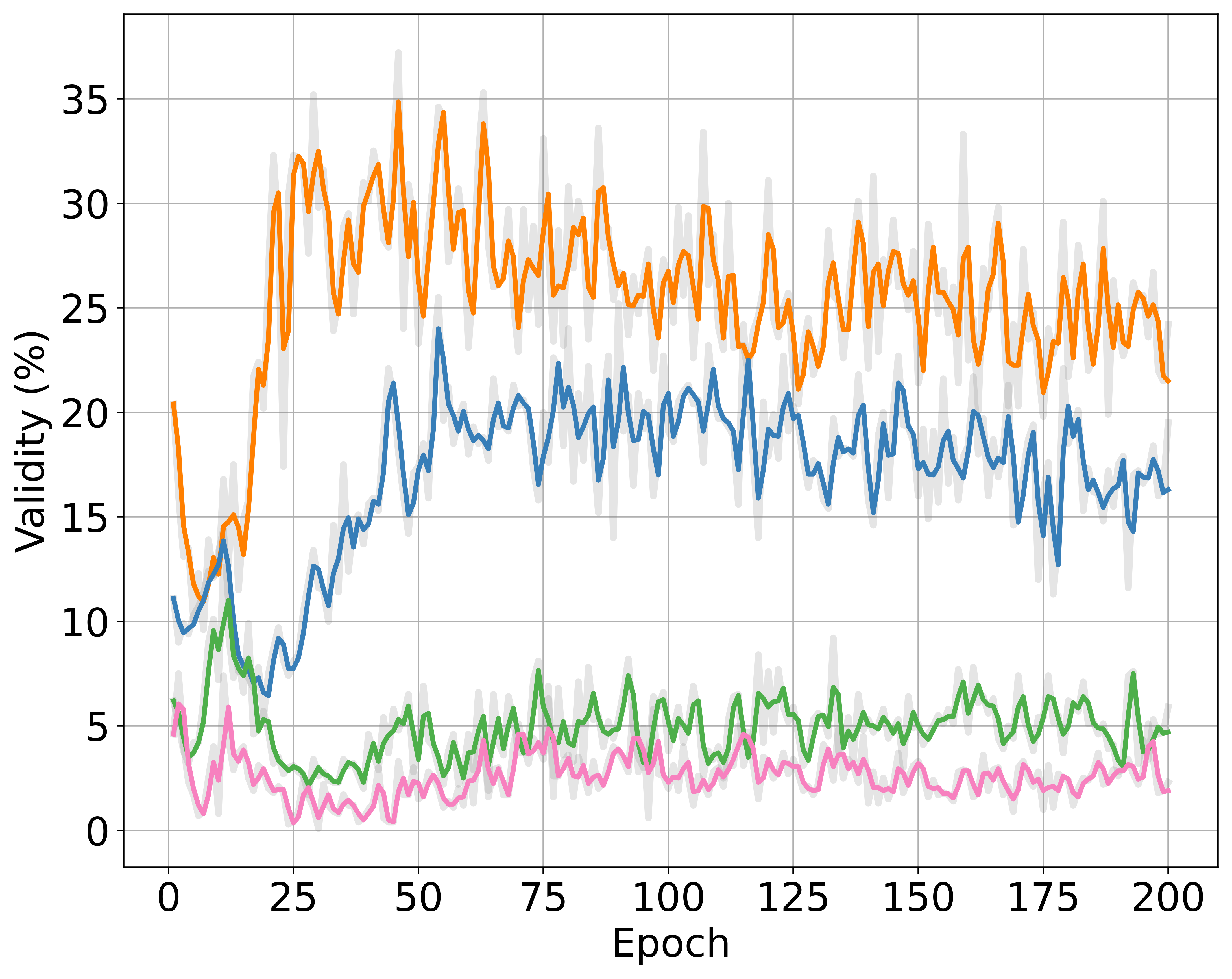} 
      \label{fig:sigma_sMMD_validity_comparison}
    \end{subfigure}
    \quad
    \begin{subfigure}{0.45\textwidth}
      \caption{}
      \centering
      \includegraphics[width=0.9\textwidth]{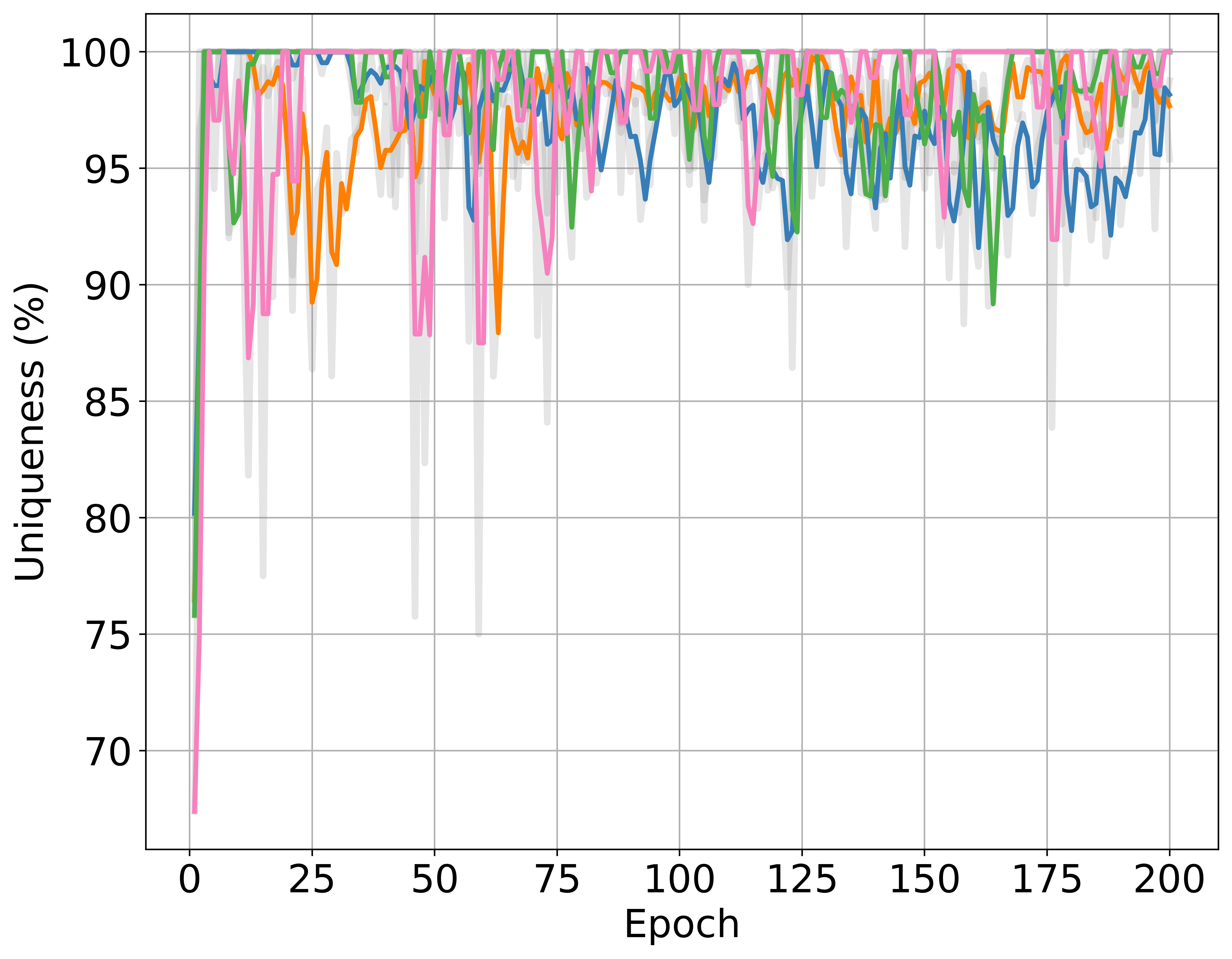}  
      \label{fig:sigma_sMMD_uniqueness_comparison}
    \end{subfigure}

    \begin{subfigure}{0.45\textwidth}
      \caption{}
      \centering
      \includegraphics[width=0.9\textwidth]{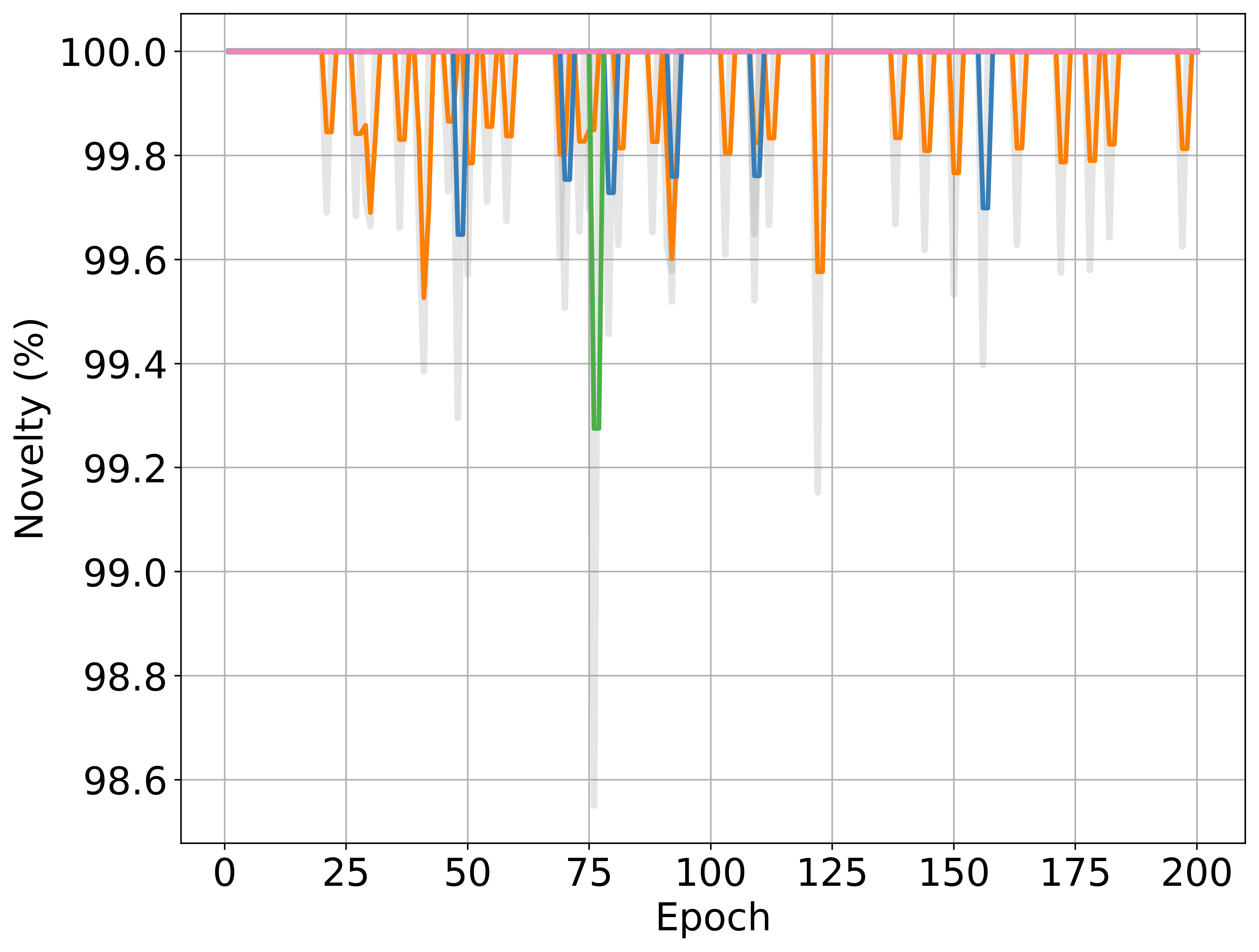}  
      \label{fig:sigma_sMMD_novelty_comparison}
    \end{subfigure}
    \quad
    \begin{subfigure}{0.45\textwidth}
      \caption{}
      \centering
      \includegraphics[width=0.9\textwidth]{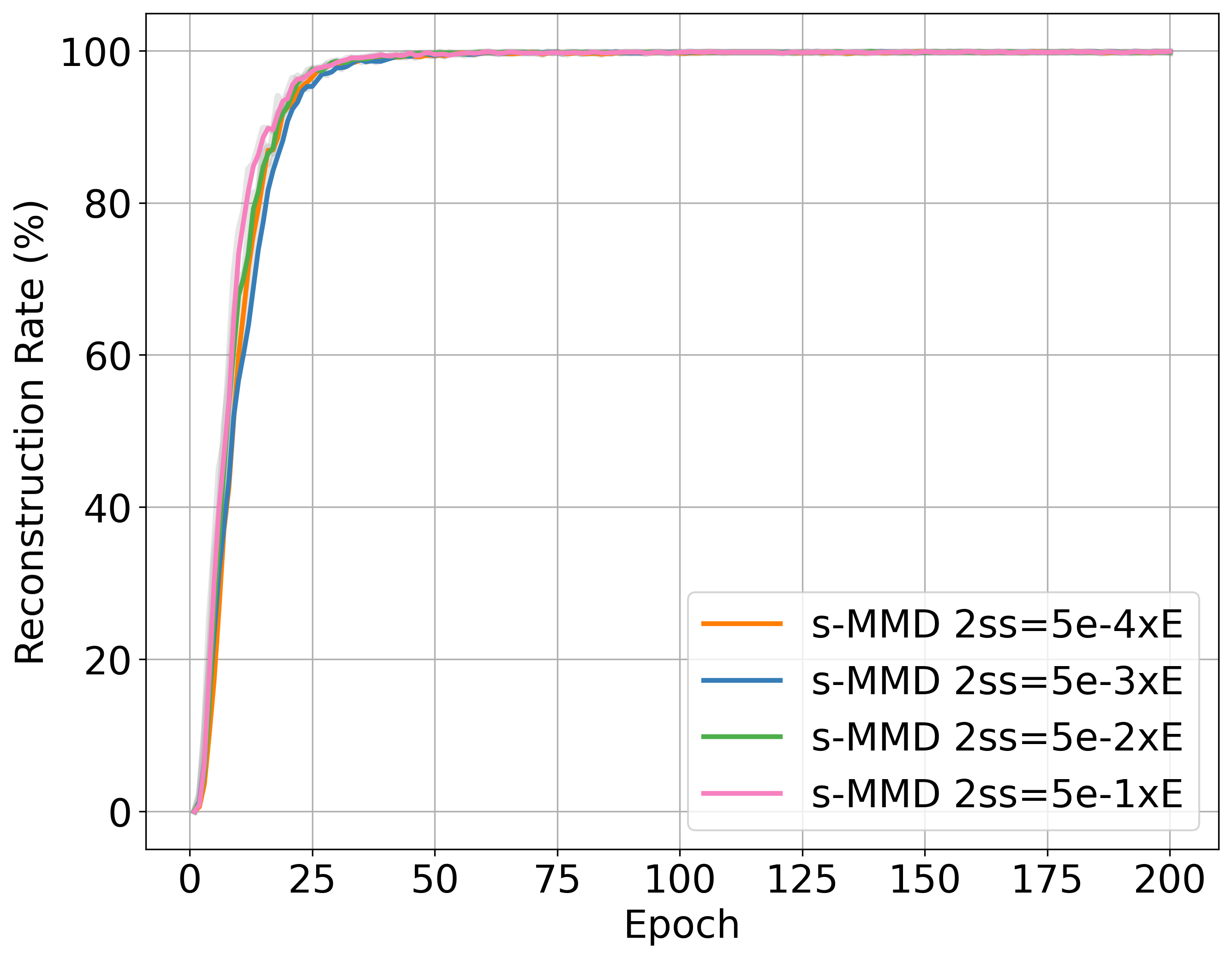} 
      \label{fig:sigma_sMMD_reconstruction_comparison}
    \end{subfigure}

\caption{\textbf{Performance comparison of the models trained with different sigma values using standard MMD loss.} (a) Validity evaluated at every epoch. (b) Uniqueness evaluated at every epoch. (c) Novelty evaluated at every epoch. (d) Reconstruction rate evaluated at every epoch. Note that 2ss (2 sigma squared) in the legend represents the value used for $2\sigma^2$ in Eq.~\ref{eq:Kernel} and $E$ is the embedding dimension. Validity, uniqueness, and novelty are calculated at the end of each epochs using 1000 randomly generated molecules from each of the models. And the reconstruction rate is calculated using 1000 molecules from the validation set.}

\label{fig:sigma_sMMD_comparison}
\end{figure*}

In Figure ~\ref{fig:sigma_mMMD_comparison} and Figure ~\ref{fig:sigma_sMMD_comparison}, we compare the model performance of different sigma values of the kernel (Eq.~\ref{eq:Kernel}). It can be observed that the final uniqueness, novelty, and reconstruction rate are similar, while there are clear differences in validity performance. Therefore, the sigma value that gives the highest final validity rate is considered optimal. It can be observed that lower $2\sigma^2$ values give higher validity rates and $2\sigma^2 = 0.0005 \times E$ is the optimal value for both m-MMD and s-MMD models. At the optimal sigma value, the m-MMD model has higher validity rate than the s-MMD model. Besides, if models are trained with even lower sigma values ($2\sigma^2 = 0.00005 \times E$ for example), the models would break down because they cannot get gradient information from the MMD loss term (results not shown).

\section{$\delta$ Comparison}
\begin{figure*}
    \centering
    \captionsetup[subfigure]{position=top, labelfont=bf, textfont=normalfont, singlelinecheck=off, justification=raggedright}
    \begin{subfigure}{0.45\textwidth}
      \caption{}
      \centering
      \includegraphics[width=0.9\textwidth]{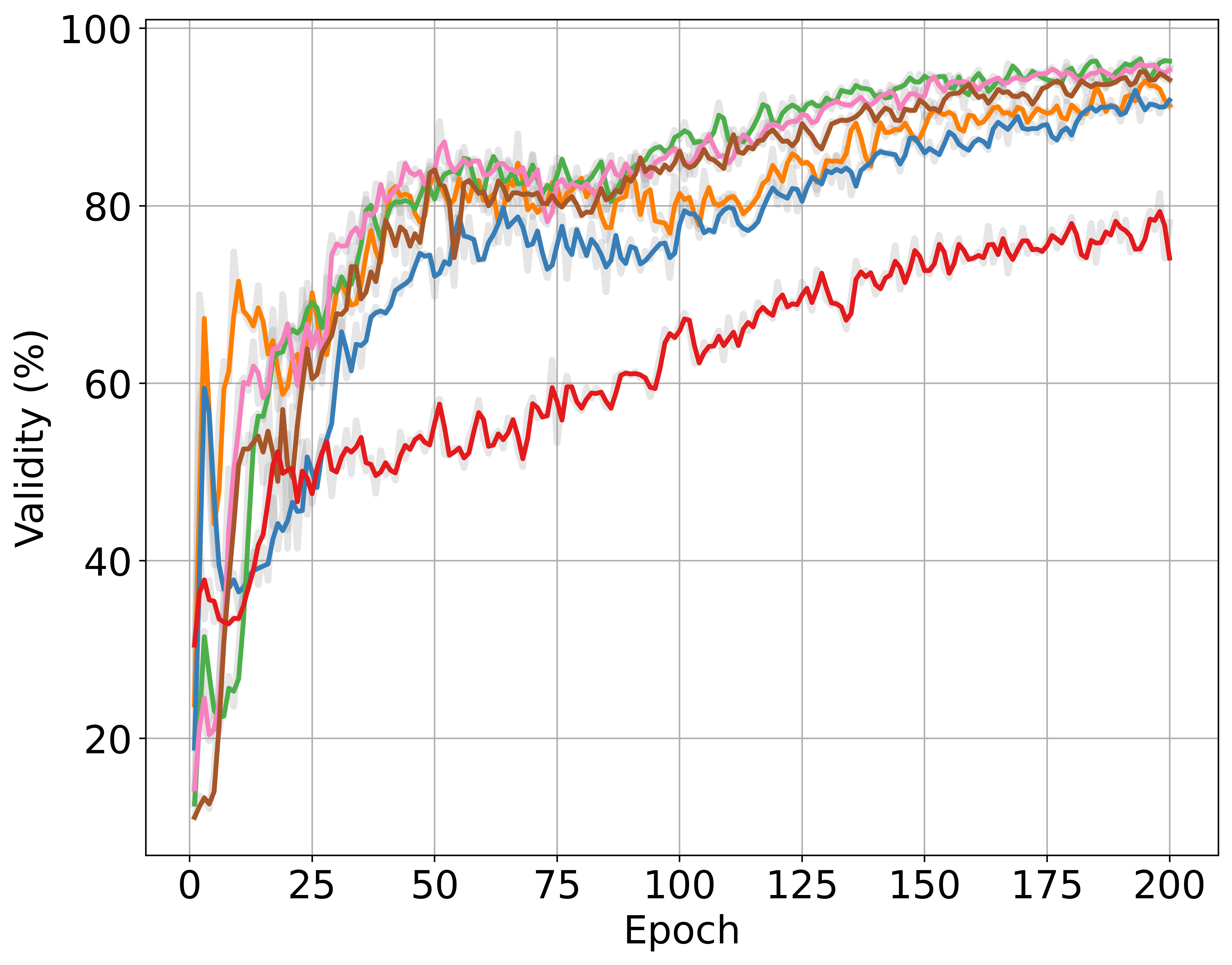} 
      \label{fig:delta_validity_comparison}
    \end{subfigure}
    \quad
    \begin{subfigure}{0.45\textwidth}
      \caption{}
      \centering
      \includegraphics[width=0.9\textwidth]{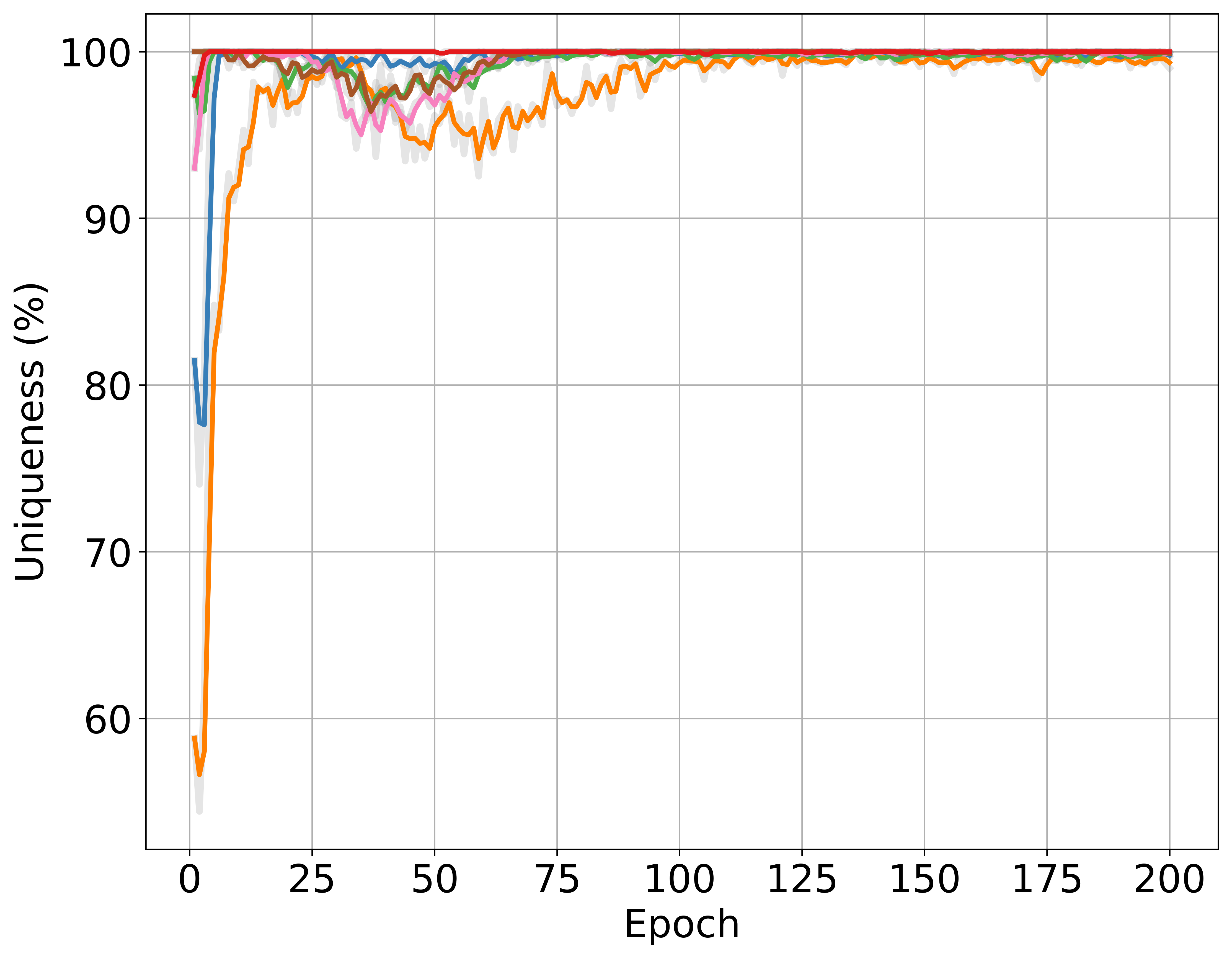}  
      \label{fig:delta_uniqueness_comparison}
    \end{subfigure}

    \begin{subfigure}{0.45\textwidth}
      \caption{}
      \centering
      \includegraphics[width=0.9\textwidth]{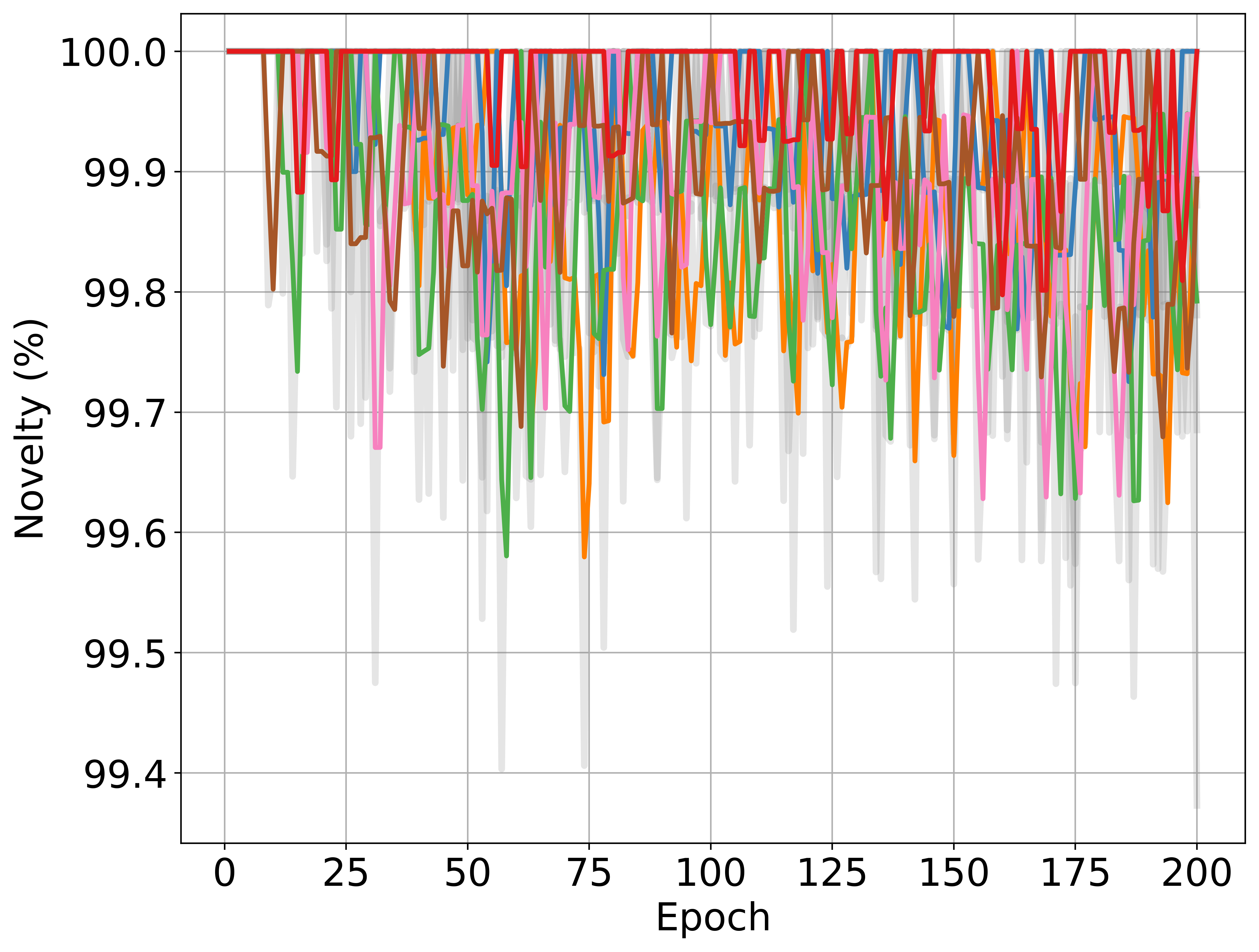}  
      \label{fig:delta_novelty_comparison}
    \end{subfigure}
    \quad
    \begin{subfigure}{0.45\textwidth}
      \caption{}
      \centering
      \includegraphics[width=0.9\textwidth]{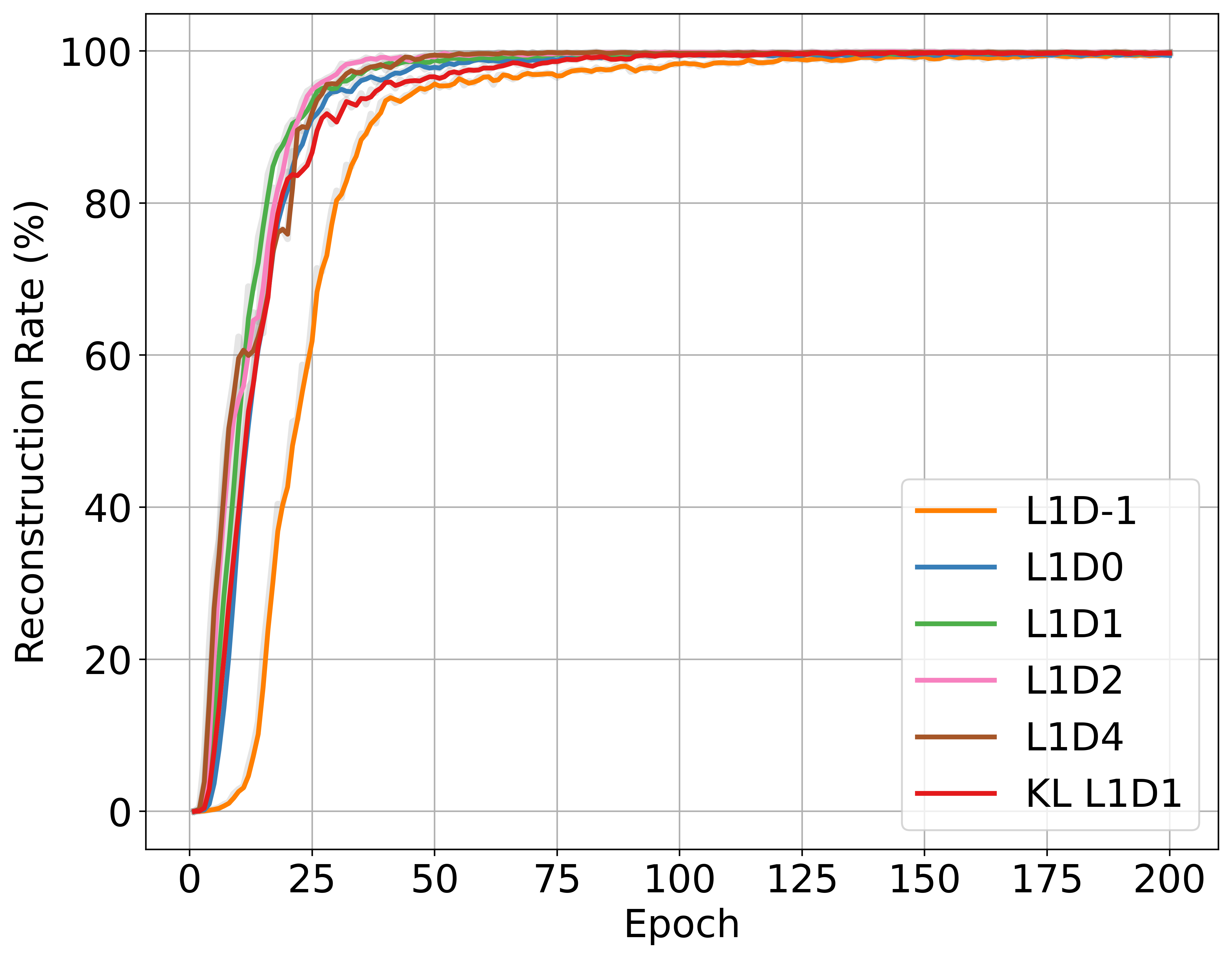} 
      \label{fig:delta_reconstruction_comparison}
    \end{subfigure}

\caption{\textbf{Performance comparison of the models trained with different $\delta$ values (with $\lambda=1$) using modified MMD loss and KL loss.} (a) Validity evaluated at every epoch. (b) Uniqueness evaluated at every epoch. (c) Novelty evaluated at every epoch. (d) Reconstruction rate evaluated at every epoch. Validity, uniqueness, and novelty are calculated at the end of each epoch using 1000 randomly generated molecules from each of the models. And the reconstruction rate is calculated using 1000 molecules from the validation set. Note that LxDy in the legend means that the model is trained with $lambda=x$ and $delta=y$. For example, the model labeled with L1D-1 is trained with $\mathcal{L}(\lambda=1, \delta=-1)$}
\label{fig:delta_comparison}
\end{figure*}

We designed the $\delta$ such that when $\lambda$ is 1, and $\delta$ is greater than -1, the AE-like term contributes to the model reconstruction performance. When $\delta$ is large, the model ignores the regions with added noise and thus is turned into a pure auto-encoder. When $\delta$ is equal to -1, the model is VAE-like where each latent vector is treated as a distribution. When $\delta$ is in between these two extrema, the model achieves the AE-like reconstruction rate while obtaining better generative performance in NUV metrics (Figure ~\ref{fig:delta_comparison}).

\section{KAE Interpolation}
\begin{figure}[!h]
    \centering
    \includegraphics[width=0.8\textwidth]{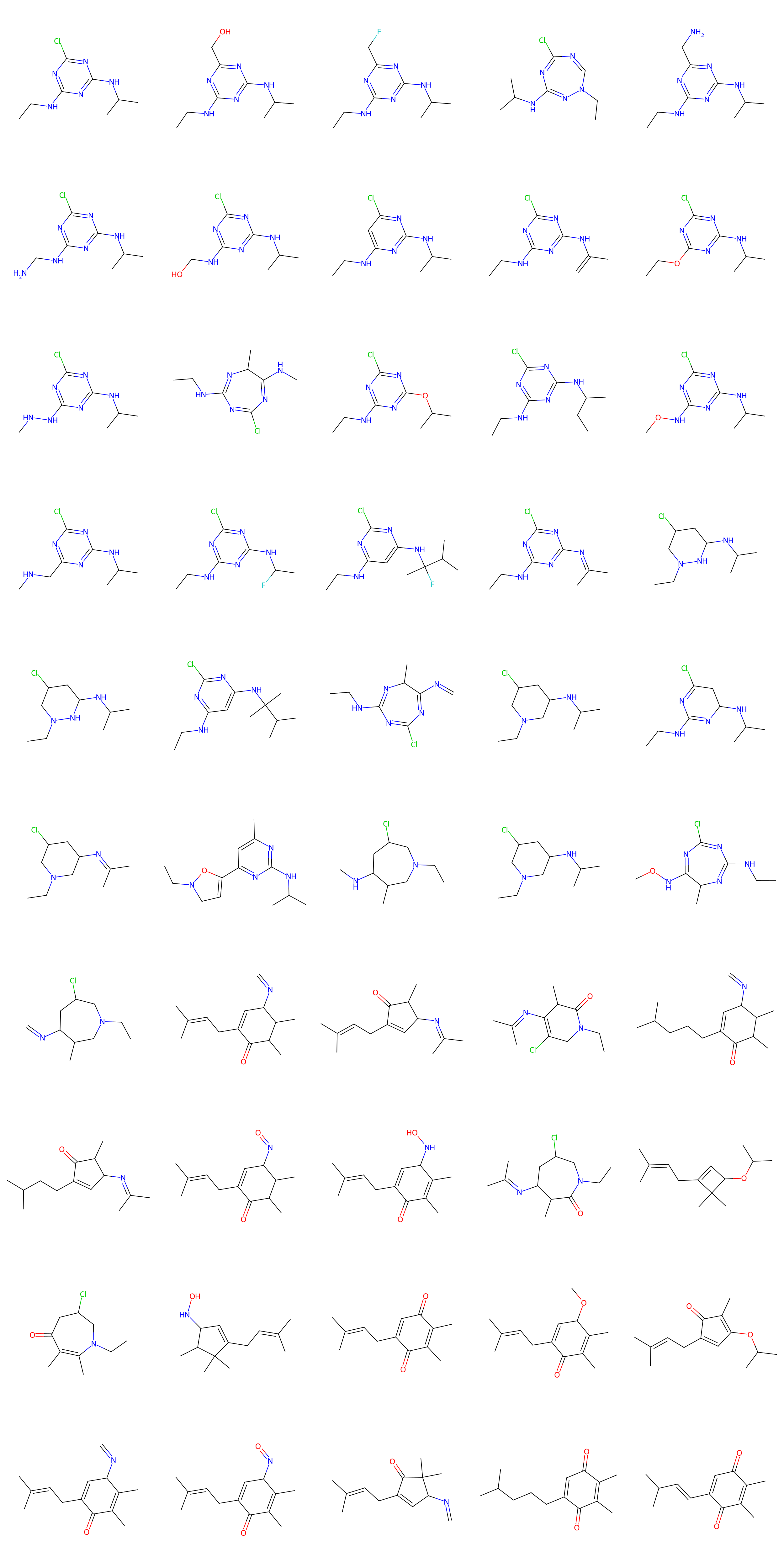}
    \caption{\textbf{Interpolation of KAE Latent Space} Atrazine and Plastoquinone are encoded into two latent vectors. The model then decodes from Atrazine to Platsoquinone in 100 evenly spaced steps with a beam size of 30.}
    \label{fig:KAE_Interpolation}
\end{figure}

KAE has near 100\% reconstruction rate and this gives it an advantage to perform modifications precisely around the input molecules. We show an example by linearly interpolating from Atrazine to Plastoquinone. The two molecules were encoded into their latent representations and the model decoded Atrazine to the Plastoquinone in 100 evenly spaced steps with a beam size of 30. The NUV molecules along this trajectory is plotted in Figure~\ref{fig:KAE_Interpolation}. The result shows an effective mixing/transitioning of the two molecules' motifs and exact reconstructions of the starting and ending compounds.

\end{document}